\documentclass{article} % For LaTeX2e
 
\PassOptionsToPackage{table,dvipsnames}{xcolor}
 
\usepackage[final]{colm2026_conference}
 
\usepackage{microtype}
\usepackage{hyperref}
\usepackage{url}
\usepackage{booktabs}
\usepackage{multirow}
\usepackage{adjustbox}
\usepackage{graphicx}
\usepackage{listings}
\usepackage[table,dvipsnames]{xcolor}
\usepackage{colortbl}
\usepackage{amssymb}
\usepackage{amsmath}
\usepackage{float}
\usepackage{enumitem}
\usepackage{pifont}
\usepackage{tcolorbox}
\tcbuselibrary{skins}
\usepackage{etoolbox}
\usepackage{expl3}
\usepackage[T1]{fontenc}
 
\newenvironment{reusetable}[1][!htbp]{\begin{table}[#1]}{\end{table}}
 
\definecolor{lightgray}{gray}{0.9}
 
\AtBeginEnvironment{figure}{\setlength{\abovecaptionskip}{-6pt}}
\AtBeginEnvironment{table}{\setlength{\abovecaptionskip}{6pt}}
\definecolor{posgreen}{HTML}{1A7F37}
\definecolor{negred}{HTML}{CF222E}

\newcommand{\spspace}{\textvisiblespace{}\hspace{0.08em}}
\newcommand{\spnl}{\texttt{\textbackslash n}}
 
\newcommand{\tok}[1]{%
  \tcbox[on line, arc=1.5pt, outer arc=1.5pt,
    boxsep=0pt, left=2pt, right=2pt, top=2pt, bottom=2pt,
    boxrule=0.3pt, colframe=gray!50, colback=gray!8,
    fontupper=\ttfamily\footnotesize]{\strut#1}}
 
\ExplSyntaxOn
\NewDocumentCommand{\tokens}{m}{
  \clist_set:Nn \l_tmpa_clist { #1 }
  \int_zero:N \l_tmpa_int
  \clist_map_inline:Nn \l_tmpa_clist {
    \int_incr:N \l_tmpa_int
    \int_compare:nNnT { \l_tmpa_int } > { 1 } { \, }
    \tok{##1}
  }
}
\ExplSyntaxOff
 
\usepackage{lineno}
 
\definecolor{darkblue}{rgb}{0, 0, 0.5}
\hypersetup{colorlinks=true, citecolor=darkblue, linkcolor=darkblue, urlcolor=darkblue}
 
\title{Shorthand for Thought: Compressing LLM Reasoning via Entropy-Guided Supertokens}
 
\author{Zhenyu Zhao \\
Lila Sciences\thanks{Work done at Writer, Inc.} \\
\texttt{zhenyuzhao@lila.ai} \\
\And
Sander Land \\
Writer, Inc. \\
\texttt{sander@writer.com} \\
\And
Dan Bikel \\
Writer, Inc. \\
\texttt{dan@writer.com} \\
\And
Waseem Alshikh \\
Writer, Inc. \\
\texttt{waseem@writer.com} \\
}
 
\begin{document}
 
\ifcolmsubmission
\linenumbers
\fi
 
\maketitle
 
\begin{abstract}
Reasoning in Large Language Models incurs significant inference-time compute, yet the token-level information structure of reasoning traces remains underexplored. We observe that reasoning tokens split into two functional types: low-entropy \textit{structural} tokens (recurring phrases that scaffold the reasoning process) and higher-entropy \textit{organic} tokens (problem-specific content that drives toward a solution). This asymmetry motivates a simple, model-agnostic compression pipeline: apply cross-word BPE merges on a model's own reasoning traces to derive \textit{supertokens} that capture frequent structural patterns, then teach the model to adopt them via supervised fine-tuning. Across three model families and five mathematical reasoning benchmarks, our approach shortens reasoning traces by 8.1\% on average; under a TOST equivalence analysis at a $\pm 2$~pp margin, accuracy is equivalent or inconclusive on 14/15 model--benchmark cells (3 pass equivalence, 11 inconclusive, predominantly AIME at $N{=}30$), with non-equivalent degradation on 1/15 cells (DeepSeek-R1-Distill-Llama-70B on MATH-500). Beyond compression, learned supertokens often align with interpretable reasoning moves such as backtracking, verification, and strategy shifts. This enables a compact structural analysis of reasoning traces: correct traces show more recovery and verification patterns, while incorrect traces show more repeated hedging and unresolved counterarguments.
We release the full pipeline as open-source code.\footnote{Code available at \url{https://github.com/Writer/shorthand-for-thought}.}
\end{abstract}
 
\section{Introduction}
\label{sec:intro}
 
Large Language Models (LLMs) have achieved strong performance gains through chain-of-thought (CoT) reasoning, generating extended thinking tokens before producing answers~\citep{wei2022chain, openai2024o1, Guo_2025}. CoT reasoning has shown consistent benefits across mathematics, coding, and agentic planning~\citep{lewkowycz2022solving, chen2021codex, yao2023react}.
 
However, reasoning traces incur substantial computational cost, and efforts to compress them have focused on replacing explicit reasoning with latent states or reducing length through distillation~\citep{deng2024explicit, chen2024cot_distillation}. Comparatively little attention has been given to the \emph{content} of reasoning outputs: the linguistic structure and information density of the traces themselves.
 
In this paper, we take a first step toward modeling LLM reasoning in a more structured and principled manner. We propose that reasoning traces are not a homogeneous stream of tokens, but rather a composition of two functionally distinct token populations that we term \textbf{structural} and \textbf{organic} reasoning tokens:
 
\begin{itemize}[nosep]
    \item \textbf{Structural tokens}: recurring, formulaic phrases that scaffold reasoning, such as backtracking cues (``Wait, hold on''), verification phrases (``let us verify''), and strategy shifts (``Let me try''), which organize the reasoning flow but carry little problem-specific information.
    \item \textbf{Organic tokens}: problem-specific content that drives toward a solution, including mathematical expressions, variable bindings, intermediate results, and reasoning direction changes.
\end{itemize}
 
We formalize this distinction using information theory. By measuring per-token conditional entropy across a large corpus of CoT traces, we find that structural tokens concentrate in frequent multi-token patterns whose continuations are near-deterministic once the phrase begins, while organic tokens exhibit substantially higher entropy. We are not the first to observe this split: \citet{wang2026beyond} show that roughly 80\% of CoT tokens are generated with high confidence, and \citet{yuan2025tokensneedthinking} show that many thinking tokens contribute minimally to the final answer. Prior work uses the split to decide which tokens receive RL gradient or to prune low-importance content, in both cases holding the vocabulary fixed; we instead hold training fixed and change the vocabulary, turning the recurring low-entropy structural spans, whose information content is not commensurate with the cost of generating them one token at a time, into supertokens for lossless compression.

We exploit this by applying cross-word BPE merges~\citep{liu2025superbpe} to each model's own reasoning traces, deriving model-specific \textit{supertokens} that we append to the base vocabulary and teach the model to adopt via supervised fine-tuning (SFT) of only the embedding layer, LM head, and a few transformer layers. The pipeline is model-agnostic: (1)~collect reasoning traces, (2)~derive supertokens via BPE merges, (3)~fine-tune to adopt them.
 
This shortens reasoning traces by 8.1\% on average across five benchmarks; a TOST equivalence analysis at $\pm 2$~pp (Appendix~\ref{sec:significance-appendix}) finds accuracy equivalent or inconclusive on 14/15 model--benchmark cells (3 equivalent, 11 inconclusive, predominantly AIME at $N{=}30$), with non-equivalent degradation on 1/15 cells (DeepSeek-R1-Distill-Llama-70B on MATH-500; Appendix~\ref{sec:limitations}). Unlike content-level methods that prune steps or replace them with latent states, our vocabulary-level approach preserves the \emph{complete, readable trace}: we compress the \emph{expression} of reasoning, not its \emph{content}. Because many supertokens correspond to recognizable reasoning moves, the compressed trace reads as a sequence of structural signposts (Figure~\ref{fig:schematic-reasoning}).

Our contributions are:
\begin{enumerate}
\item A \textbf{simple, model-agnostic pipeline} for reasoning compression requiring only BPE-based supertoken derivation and supervised fine-tuning.
\item An \textbf{information-theoretic analysis} of reasoning traces via the structural/organic decomposition, showing that supertokens predominantly capture low-entropy structural patterns.
\item \textbf{8.1\% reasoning compression with accuracy equivalent or inconclusive at $\pm 2$~pp on 14/15 model--benchmark cells} (TOST; 11/15 inconclusive, predominantly AIME; 1/15 non-equivalent on DS-R1-Llama / MATH-500) across three model families and five benchmarks, with evidence that learned supertokens align with interpretable reasoning idioms, enabling a descriptive structural analysis of correct and incorrect reasoning traces.

\end{enumerate}
 
\begin{figure}[t]
  \centering
  \includegraphics[width=\linewidth]{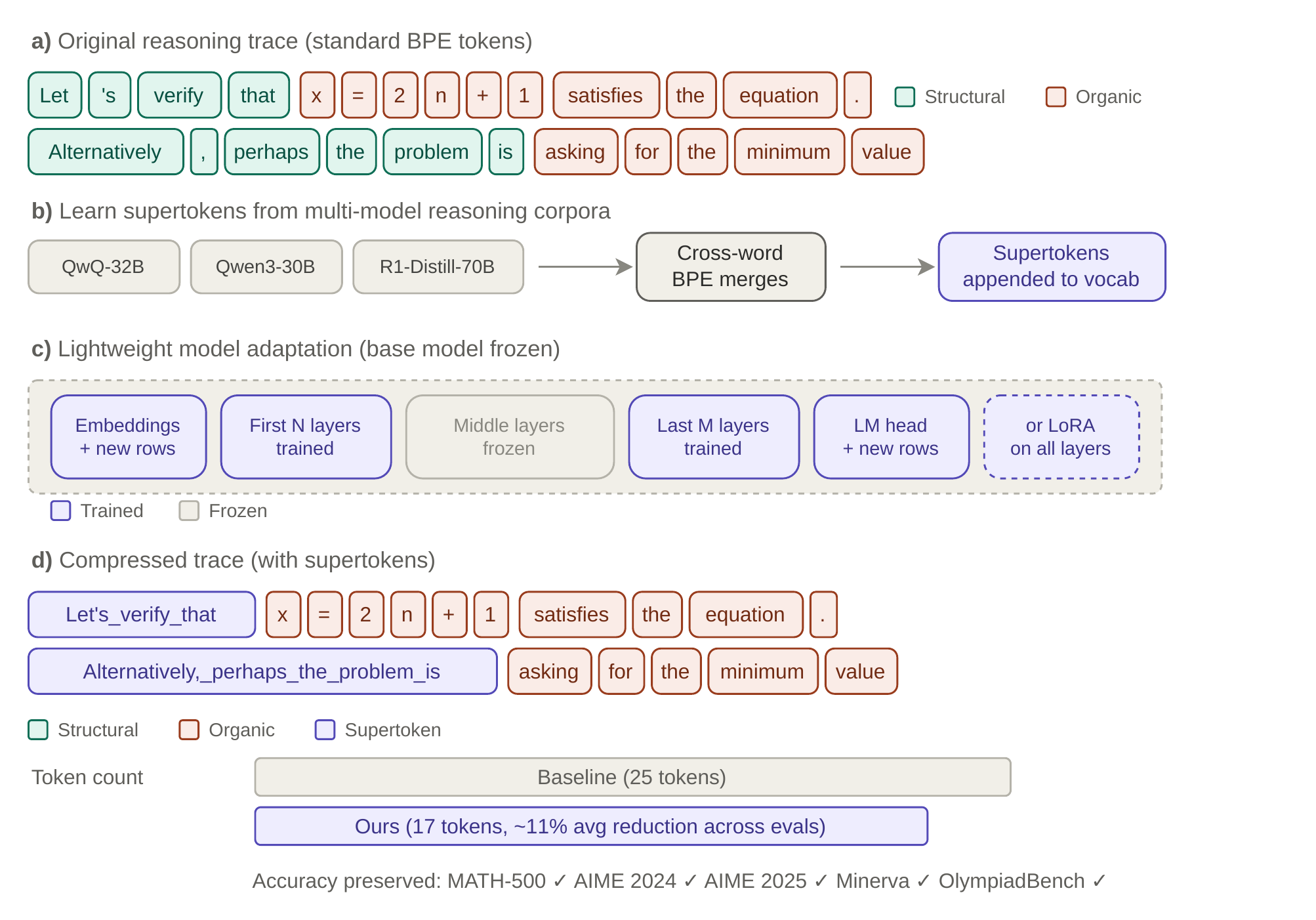}
  \caption{%
    Overview of the pipeline: (1)~derive reasoning-specific supertokens via BPE merges on reasoning traces, then (2)~adapt the model to utilize these supertokens via supervised fine-tuning to produce shorter reasoning trajectories with accuracy equivalent or inconclusive at $\pm 2$~pp on 14/15 model--benchmark cells under a TOST analysis (3 equivalent, 11 inconclusive; Appendix~\ref{sec:significance-appendix}). %
  }
  \label{fig:pipeline-overview}
\end{figure}
 
\section{Related Work and Preliminaries}
 
\paragraph{Superword tokenization.}
Standard BPE tokenizers~\citep{sennrich-etal-2016-neural, radford2019language} forbid tokens from spanning whitespace, limiting efficiency at larger vocabulary sizes. SuperBPE~\citep{liu2025superbpe} and BoundlessBPE~\citep{schmidt2025boundlessbytepairencoding} relax this constraint to learn cross-whitespace superwords, cutting tokens by up to 33\% on general text. Both target pretraining efficiency on general corpora; we instead apply the idea to reasoning traces from a specific model family, where high-frequency n-grams are structurally distinct and amenable to targeted filtering. As a drop-in tokenizer change, our method's natural reference class is the tokenizer literature, where corpus-level gains between algorithms are typically $1$--$2\%$ (PickyBPE~\citep{chizhov-etal-2024-bpe}: ${\sim}1.1\%$), so our $8.1\%$ is large; the $30$--$70\%$ content-level figures (below) lie on a different axis and are not directly comparable.

\paragraph{Vocabulary expansion without retraining.}
\citet{kaplan2025from} show that LLMs internally assemble multi-token word representations into coherent single vectors by intermediate layers, enabling vocabulary expansion without retraining core parameters. We follow this approach, initializing new supertoken embeddings by averaging their constituent token embeddings.
\paragraph{Structured reasoning.}
Several works impose meta-level structure on the reasoning process~\citep{yao2023react,Besta_2024,yao-etal-2024-got}; e.g., PLAN-AND-BUDGET~\citep{lin2026planbudgeteffectiveefficient} allocates a complexity-based budget over structured sub-questions and \citet{tan-etal-2025-enhancing-logical} adds symbolic logic scaffolds. These modify the reasoning \emph{process}; we instead compress its \emph{output} at the vocabulary level.
 
\paragraph{Interpreting reasoning traces.}
Recent work analyzes chain-of-thought as structured behavioral objects: \citet{gandhi2025cognitive} study cognitive behaviors (verification, backtracking, subgoal setting, backward chaining), \citet{bogdan2025thoughtanchors} identify steps that disproportionately influence later reasoning, and \citet{shahariar2025modelinghierarchicalthinkinglarge} model traces with hierarchical finite-state abstractions. Our taxonomy overlaps with these but is derived from frequent supertokens learned for compression rather than a manual annotation scheme.

\paragraph{Latent compression of reasoning traces.}
A separate line replaces discrete tokens with latent representations: Token Assorted~\citep{su2025token} uses VQ-VAE latent tokens for initial steps and Coconut~\citep{hao2025training} reasons in continuous latent space. These achieve strong compression but require specialized training objectives and lose interpretability.
 
\paragraph{Explicit CoT compression.}
A growing body of work compresses chain-of-thought at the content level. Token- and chunk-level pruning or rewriting methods, including TokenSkip~\citep{xia-etal-2025-tokenskip}, R1-Compress~\citep{wang2025r1compresslongchainofthoughtcompression}, ConMax~\citep{hu2026conmaxconfidencemaximizingcompressionefficient}, CtrlCoT~\citep{fan2026ctrlcotdualgranularitychainofthoughtcompression}, and Extra-CoT~\citep{tang2026towards}, report $20$--$73\%$ reductions, but largely on short-CoT, instruction-tuned backbones (GSM8K, MATH-500); on long-CoT traces the accuracy cost grows sharply, with independent re-evaluations showing TokenSkip incurring $>$20\,pp drops at comparable compression~\citep{hu2026conmaxconfidencemaximizingcompressionefficient, fan2026ctrlcotdualgranularitychainofthoughtcompression}. A related family controls length through training rather than per-token pruning: O1-Pruner~\citep{luo-etal-2026-o1}, CoT-Valve~\citep{ma-etal-2025-cot}, and AutoL2S~\citep{luo2026autol2sautolongshortreasoning} change \emph{what} the model produces and rely on RL, a learned controller, or a switching mechanism. All operate at the content level: they remove, shorten, or hide explicit reasoning steps, creating a genuine efficiency--interpretability tradeoff, and are typically tuned in-distribution per benchmark. Our approach instead makes a one-time, out-of-distribution vocabulary change applied as a drop-in tokenizer, leaving the complete trace readable and grammatical (which matters for debugging and oversight), and we evaluate on long-CoT reasoning models (QwQ-32B, Qwen3-30B-A3B, DeepSeek-R1-Distill-Llama-70B) over AIME'24, AIME'25, MATH-500, Minerva, and Olympiad, where baseline traces span thousands of tokens. Table~\ref{tab:method-comparison} positions our approach relative to these and the latent methods above.

\begin{table}[t]
\centering
\small
\renewcommand{\arraystretch}{1.15}
\begin{tabular}{@{}llccc@{}}
\toprule
\textbf{Method} & \textbf{Approach} & \textbf{Token $\downarrow$} & \textbf{$\Delta$Acc} & \textbf{Interpret.} \\
\midrule
TokenSkip        & Step pruning       & 30--40\%  & $<$4\% drop  & \ding{55} \\
R1-Compress      & Chunk compression  & $\sim$20\% & 0.6\% drop  & \ding{55} \\
ConMax           & RL compression     & 43\%       & 0.7\% drop  & \ding{55} \\
CtrlCoT          & Dual-granularity   & 31\%       & +7.6 pp$^*$ & \ding{55} \\
Extra-CoT        & Extreme ratio      & $>$73\%    & +0.6 pp$^\dagger$ & \ding{55} \\
Coconut          & Latent space       & High       & Varies       & \ding{55} \\
Token Assorted   & VQ-VAE latent      & High       & Varies       & \ding{55} \\
\midrule
\textbf{Ours}    & \textbf{Vocabulary} & \textbf{8.1\%} & \textbf{$\sim$0 pp}$^{\ddagger}$ & \ding{51} \\
\bottomrule
\end{tabular}
\caption{Comparison with CoT compression methods. Content-level and latent methods achieve higher compression but alter or remove reasoning steps.
Our vocabulary-level approach preserves the complete readable trace, unlike methods that remove, summarize, or hide reasoning steps. $^*$CtrlCoT's +7.6\,pp is relative to the TokenSkip baseline on MATH-500 (not the uncompressed model). $^\dagger$Extra-CoT's +0.6\,pp is relative to the uncompressed model on MATH-500 with Qwen3-1.7B. $^{\ddagger}$Pooled aggregate across 15 model--benchmark cells; per-cell breakdown via TOST at $\pm 2$~pp in Table~\ref{tab:tost} (3/15 equivalent, 11/15 inconclusive, 1/15 non-equivalent on DeepSeek-R1-Distill-Llama-70B / MATH-500).}
\label{tab:method-comparison}
\end{table}
 
\section{Information-Theoretic Motivation}
\label{sec:entropy-motivation}
We now formalize the compression opportunity suggested by the structural/organic distinction, drawing on Shannon's information-theoretic framework~\citep{shannon1948mathematical, shannon1951prediction}. The key observation is that structural tokens concentrate in frequent \emph{multi-token patterns} whose continuation tokens are highly predictable once the phrase begins: once a model commits to \textit{``Let me reconsider''}, the remaining tokens of the phrase carry almost no information. We quantify this sequential redundancy via conditional entropy. Given a reasoning trace $T = (t_1, \ldots, t_n)$ generated by a model with parameters $\theta$, the conditional entropy at position $i$ is:
\begin{equation}
  H_\theta(t_i \mid t_{<i}) = -\sum_{v \in \mathcal{V}} P_\theta(t_i = v
  \mid t_{<i}) \log P_\theta(t_i = v \mid t_{<i})
\end{equation}
Let $\mathcal{M} \subset \{1, \ldots, n\}$ denote positions within a multi-token merge span, with mean entropy $h_\mathcal{M} = \frac{1}{|\mathcal{M}|} \sum_{i \in \mathcal{M}} H_\theta(t_i \mid t_{<i})$. If a fraction $\rho = |\mathcal{M}|/n$ of tokens participate in merges, the theoretical compression ceiling is:
\begin{equation}
  \Delta = \rho \cdot \left(1 - \frac{h_\mathcal{M}}{\log_2 |\mathcal{V}|}
  \right)
  \label{eq:compression_ceiling}
\end{equation}
When $h_\mathcal{M} \ll \log_2|\mathcal{V}|$, this simplifies to $\Delta \approx \rho$: nearly all tokens within merge spans can be absorbed, so a tokenization algorithm learning frequent patterns over reasoning traces should preferentially discover structural sequences.
 
\section{Learning supertokens from reasoning traces}
\subsection{Vocabulary construction via SuperBPE}
\label{sec:vocab}
We apply SuperBPE~\citep{liu2025superbpe} to a corpus of
reasoning traces to derive a small set of \emph{reasoning supertokens} that
are appended to the base vocabulary of a pretrained LLM.
The key design choices are (1) which token n-grams to consider as supertoken candidates,
and (2) how to initialize the new token embeddings without retraining the
transformer body.
\paragraph{Corpus and n-gram extraction.}
We generate reasoning traces from each target model on OpenThoughts3~\citep{guha2026openthoughts} prompts; each model family receives its own supertokens from its own traces. The entropy analysis in Section~\ref{sec:entropy-validation} uses QwQ-32B~\citep{qwen2025qwq32b} as the representative model; cross-model generality is validated in Appendix~\ref{sec:crossmodel-entropy}.
N-gram counts are computed over reasoning traces only, with per-sample counts capped at 10 to prevent high-repetition samples from dominating merge selection~\citep{land2024fishingmagikarpautomaticallydetecting}.
 
\paragraph{Filter design.}
Unfiltered BPE on reasoning text produces merges spanning syntactic boundaries (punctuation-initial sequences, cross-clause concatenations, multi-digit runs) unlikely to generalize.
We define a \emph{structural filter} restricting merges to four surface patterns: capitalized phrase-initial spans (e.g.\ \textit{``The answer is''}), punctuation-plus-newline continuations, comma-led continuations, and space-prefixed single digits (multi-digit concatenations excluded).
This reduces raw compression from ${\sim}$28\% to ${\sim}$10\% at 1000 merges (Figure~\ref{fig:merge-compression}, Appendix), producing semantically coherent reasoning phrases (Table~\ref{tab:token-examples}, Appendix).
 
The four rules are a fixed surface-form predicate with no numeric hyperparameters and were specified independently of any downstream evaluation. To confirm they generalize beyond the trace slice used to inspect candidate n-grams, we set aside the last 20 of 120 OpenThoughts3-1.2M shards (200{,}000 traces, $1.01$B reasoning tokens) as a held-out slice and re-derived filtered BPE merges on this slice using the identical filter. The held-out-derived vocabulary (954 merges) saves $7.64\%$ of held-out reasoning tokens at 1000 merges, indistinguishable from the production vocabulary (757 merges) at $7.36\%$ on the same data; $80.6\%$ of the production top-100 merged phrases also appear in the held-out top-100. No retraining or new downstream evaluations were performed for this check.

\paragraph{Merge budget.}
After filtering, we retain the top-$K{=}250$ merges ranked by corpus frequency, each introducing one new vocabulary entry (e.g., 151{,}669 $\to$ 151{,}919 for the Qwen-family models). Table~\ref{tab:token-examples} and Figure~\ref{fig:merge-compression} (Appendix) show representative statistics.
 
\paragraph{Embedding initialization.}
New token embeddings are initialized by averaging the
embeddings of their constituent tokens, providing a stable starting point
that requires no additional forward passes. This mean-of-constituents
initialization is important in practice: a random initialization of the new
supertoken rows under the same training budget failed to train stably and did
not converge to usable supertoken adoption, so we did not pursue it further.
We report this qualitatively rather than as a controlled ablation.
\paragraph{Tokenizer application.}
At both training and inference time, the base tokenizer is applied first, and
supertoken merges are applied as a post-processing step over the resulting
token sequence. This keeps the extended tokenizer compatible with the original
model and avoids modifying tokenizer internals.
\subsection{Entropy Validation of Supertokens}
\label{sec:entropy-validation}
We now test the prediction from Section~\ref{sec:entropy-motivation}: do the
merges discovered by our supertokenization algorithm capture sequential redundancy in reasoning
traces? We compute per-token conditional entropy across 10{,}000 traces
(115M tokens) from OpenThoughts3-1.2M using QwQ-32B as both generator and
scorer, align each base token with the supertoken-based tokenization via character
offsets, and classify tokens into three roles: \textbf{non-merged},
\textbf{first} (the opening token of a merge span), and
\textbf{continuation} (subsequent tokens completing the phrase).
 
\begin{figure}[t]
  \centering
  \includegraphics[width=\linewidth]{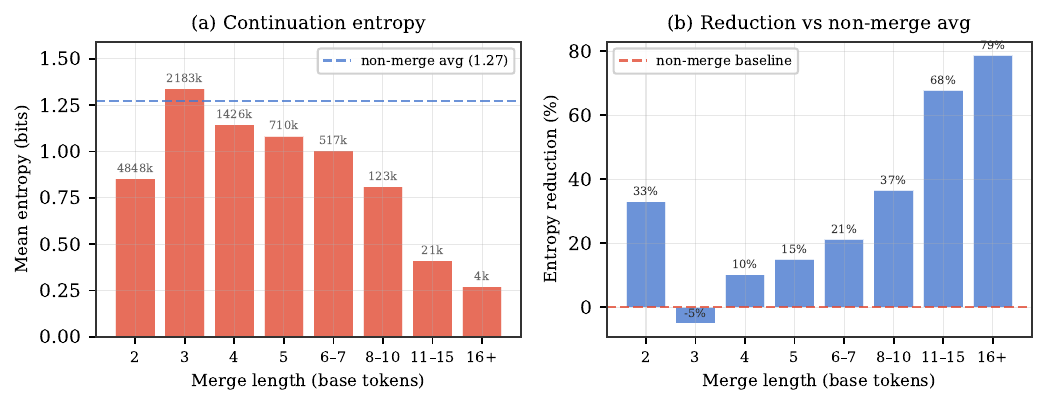}
  \caption{Continuation-token entropy by supertoken merge length.
  (a)~Mean conditional entropy per bin (dashed line: non-merge average, 1.27~bits). The gap widens from 33\% at length~2 to 79\% at length~16+; length-3 merges are an outlier ($-5\%$).
  (b)~Relative entropy reduction ($1 - \bar{H}_{\mathrm{cont}} / \bar{H}_{\mathrm{non\text{-}merge}}$); longer merges capture progressively more deterministic continuations.}
  \label{fig:merge-length}
\end{figure}
 
\paragraph{Merge length scaling.}
Our supertoken merges cover $\rho = 15.2\%$ of all token positions. As shown in Figure~\ref{fig:merge-length}, continuation-token entropy decreases with merge length: length-2 merges already show a 33\% entropy reduction relative to non-merge tokens, and merges of length $\geq$16 reach 79\% reduction. This confirms that the merges target \emph{sequential} redundancy, i.e., phrases where continuations become near-deterministic once the opening token is observed (see also Figure~\ref{fig:entropy-heatmap}, Appendix). Length-3 merges are an exception, displaying entropy on par with non-merge tokens, likely because they contain a high concentration of transitioning tokens~\citep{wang2026beyond}.
 
% eval_table.tex — include via \input{eval_table} from your main .tex file
% Requires in preamble: booktabs, multirow, xcolor (for coloring deltas)
\definecolor{darkgreen}{rgb}{0.0, 0.5, 0.0}
\definecolor{darkred}{rgb}{0.7, 0.0, 0.0}
 
\begin{reusetable}[t]
\centering
\begin{adjustbox}{max width=\textwidth}
\renewcommand{\arraystretch}{1.2}
\begin{tabular}{l cc cc ccc cc}
\toprule
& \multicolumn{2}{c}{\textbf{Baseline}}
& \multicolumn{2}{c}{\textbf{SFT (no supertoken)}}
& \multicolumn{3}{c}{\textbf{SFT (supertoken)}}
& \multicolumn{2}{c}{\textbf{Improvement}} \\
\cmidrule(lr){2-3} \cmidrule(lr){4-5} \cmidrule(lr){6-8} \cmidrule(lr){9-10}
\textbf{Benchmark}
& Acc (\%) & Avg.\ Tok.\ ($\downarrow$)
& Acc (\%) & Avg.\ Tok.\ ($\downarrow$)
& Acc (\%) & Avg.\ Tok.\ ($\downarrow$) & SuperBPE (\%)
& {$\Delta$Acc (\%)} & {$\Delta$Avg.\ Tok.\ (\%) ($\downarrow$)} \\
\midrule
 
%% ---- qwq-32b-thinking ----
\rowcolor{lightgray} \multicolumn{10}{c}{\textbf{QWQ-32B}} \\
\midrule
AIME'24   & 77.50 & 14082 & 76.9 & 14580 & 77.7 & 13160 & 7.6 & +0.2 & \color{darkgreen}-6.5   \\
AIME'25   & 69.20 & 16651 & 68.3 & 16838 & 72.7 & 14880 & 7.7 & \color{darkgreen}+3.5 & \color{darkgreen}\textbf{-10.6}$^{*}$  \\
MATH-500  & 80.60 & 4441  & 81.0 & 4421 & 80.5 &  4183 & 10.7 & -0.1 & \color{darkgreen}$-5.8^{*}$\\
Minerva   & 34.0  & 5470  & 32.6 & 5695 & 33.5 & 5418 & 8.15 & -0.5 & -1.1 \\
Olympiad  & 55.8  & 9584  & 56.7 & 9517 & 56.5 & 9256 & 11.7 & +0.7  & \color{darkgreen}$-3.0^{*}$ \\
\midrule
\textbf{Average} & 63.4 & 10046 & 63.1 & 10210 & 64.2  & 9379 & 9.17 & +0.8 & \color{darkgreen}\textbf{-6.6}$^{*}$  \\
 
%% ---- Model-2 ----
\midrule
\rowcolor{lightgray}\multicolumn{10}{c}{\textbf{QWEN-3-30B-A3B}} \\
\midrule
AIME'24   & 91.3 & 16305 & 91.9 & 16125 & 90.3 & 15246 & 10.7 & -1.0 & \color{darkgreen}-6.5  \\
AIME'25   & 84.0 & 19491 & 87.0 & 19275 & 84.1 & 18517 & 11.1 & +0.1 & - 5.0  \\
MATH-500  & 85.7 & 5488  & 85.8 & 5454 & 86.8 & 4849 & 12.9 & +1.1 & \color{darkgreen}\textbf{-11.6}$^{*}$ \\
Minerva   & 37.4 & 3595  & 38.2 & 3566 & 37.0 & 3571 & 9.7 & -0.4 & -0.66   \\
Olympiad  & 65.3 & 13309 & 64.5 & 13220 & 66.8 & 12372 & 11.7 & +1.5 & \color{darkgreen}$-7.0^{*}$  \\
\midrule
\textbf{Average} & 72.7 & 11638 & 73.5 & 11528 & 73.0 & 10910 & 11.2 & +0.3 & \color{darkgreen} \textbf{-6.3}$^{*}$  \\
 
%% ---- Model-3 ----
\midrule
\rowcolor{lightgray}\multicolumn{10}{c}{\textbf{DeepSeek-R1-Llama-70B-Distill}} \\
\midrule
AIME'24   & 66.7 & 9059  & 68.2 & 9037 & 70.2 & 7519 & 12.1 & +3.5 & \color{darkgreen}\textbf{-17.0}$^{*}$  \\
AIME'25   & 54.5 & 11332 & 53.7 & 11329 & 53.4 & 9390 & 11.6 & -1.1 & \color{darkgreen}\textbf{-17.0}$^{*}$  \\
MATH-500  & 78.1 & 2701  & 75.2 & 2683 & 74.8 & 2541 & 12.2 & \color{darkred}\textbf{-3.3} & \color{darkgreen}$-5.9^{*}$ \\
Minerva   & 31.6 & 3239  & 32.5 & 3217 & 31.4 & 3126 & 9.54 & -0.2 & \color{darkgreen}$-3.5^{*}$  \\
Olympiad  & 51.0 & 6373  & 50.3 & 6282 & 49.5 & 6370 & 11.4 & -1.5 & -0.1   \\
\midrule
\textbf{Average} & 56.4 & 6541 & 56.0 & 6510 & 55.9 & 5789 & 11.4 & -0.5 & \color{darkgreen}\textbf{-11.5}$^{*}$  \\
\bottomrule
\end{tabular}
\end{adjustbox}
\caption{SFT encourages supertoken adoption, reducing reasoning trace length across three model families and five math benchmarks. The \textbf{SFT (no supertoken)} columns report a control where each model is fine-tuned on OpenThoughts3 with its original tokenizer (no vocabulary expansion), isolating the contribution of supertoken adoption: this control leaves trace length essentially unchanged, so the compression is attributable to the supertokens rather than to SFT on the data. Accuracy deltas are evaluated by a TOST equivalence test at $\pm 2$~pp (Appendix~\ref{sec:significance-appendix}, Table~\ref{tab:tost}): 3/15 cells are equivalent, 11/15 inconclusive (predominantly AIME at $N{=}30$), and 1/15 (DS-R1-Llama on MATH-500) fails equivalence with a real accuracy loss. $^{*}$~marks token reductions whose 95\% CI excludes zero ($p < 0.05$).}
\label{tab:eval-results}
\end{reusetable}
 
\section{Experiments and Results}
 
\subsection{Experimental setup}
 
\paragraph{Supertoken adoption training.} 
We fine-tune each base model on OpenThoughts3~\citep{guha2026openthoughts} with the extended tokenizer applied to assistant turns only.
We compare three SFT modes: (i)~\emph{embedding-only} (new supertoken rows in \texttt{embed\_tokens} and \texttt{lm\_head}); (ii)~\emph{LoRA}~\citep{hu2022lora} ($r{=}16$, $\alpha{=}32$ on attention projections plus embedding rows); and (iii)~\emph{layer-unfreezing} (first $N$ and last $M$ transformer layers plus embeddings). In all modes, gradient hooks zero out base-vocabulary rows so only supertoken embeddings receive updates. The layer-unfreezing configuration used for all main-table results ($N{=}3$ first, $M{=}1$ last transformer layers, with backbone learning rate $2\times10^{-4}$) is held fixed across all 15 model--benchmark cells; there is no per-cell hyperparameter tuning, and the LoRA-vs-unfreezing comparison (Table~\ref{tab:training-ablation}) is a training-intensity ablation rather than model selection. Exact values are listed in Appendix~\ref{sec:experiment-details}.

\paragraph{Evaluation.} We evaluate on five mathematical reasoning benchmarks (AIME'24, AIME'25, MATH-500~\citep{hendrycks2021measuring}, Minerva, OlympiadBench~\citep{he-etal-2024-olympiadbench}) in a zero-shot setting using vLLM~\citep{kwon2023vllm} on 2 H200 GPUs, running each evaluation 5 times and reporting mean accuracy, completion tokens, and supertoken adoption rate (fraction of reasoning tokens from the expanded vocabulary). Throughout, we measure trace length in \emph{tokens} rather than characters: each generated token is one decoder forward pass, so re-encoding the same reasoning in fewer tokens is a genuine reduction in inference compute, not merely a change in how length is counted. Wall-clock latency reductions generally track or exceed token reductions (Table~\ref{tab:latency}).
 
\subsection{Adoption of supertokens and compression of reasoning trajectories}
\label{sec:supertokensadoption}
As shown in Table~\ref{tab:eval-results}, SFT on a retokenized dataset consistently drives supertoken adoption across all three model families and five benchmarks, reducing reasoning trace length by 6--12\% on average. The reduction is attributable to the expanded vocabulary rather than to any behavioral shortening induced by fine-tuning: the \emph{SFT (no supertoken)} control, trained on the same data with identical hyperparameters but no vocabulary expansion, leaves token length essentially unchanged (${-}0.47\%$ on average, TOST-equivalent to zero), whereas adding supertokens yields the full $-8.1\%$. The model therefore expresses the same reasoning content in fewer tokens rather than producing less reasoning, consistent with the lossless, content-preserving reading of the method. The two Qwen-family models (QwQ-32B~\citep{qwen2025qwq32b}, Qwen3-30B-A3B~\citep{yang2025qwen3technicalreport}) show robust accuracy (both improve on average, by $+0.8$ and $+0.3$~pp respectively); DeepSeek-R1-Distill-Llama-70B achieves the largest compression (up to $-17\%$ on AIME) but exhibits a $-3.3$~pp accuracy drop on MATH-500 whose mechanism we do not establish in this manuscript and discuss in Appendix~\ref{sec:limitations}. We support the accuracy-preservation claim with a two one-sided test (TOST) at a single equivalence margin of $\pm 2$~pp, using a paired $t$-test on per-question accuracy averaged across the 5 evaluation runs (Appendix~\ref{sec:significance-appendix}). Of the 15 model-benchmark cells, 3/15 demonstrate accuracy equivalence at $\pm 2$~pp (QwQ-32B on MATH-500 and OlympiadBench, Qwen3-30B-A3B on MATH-500), 11/15 are inconclusive (90\% CI straddles a margin bound, predominantly AIME at $N{=}30$), and 1/15 fails equivalence with a real accuracy loss: DeepSeek-R1-Distill-Llama-70B on MATH-500 ($-3.3$~pp, 90\% CI $[-5.38, -1.22]$), discussed in Appendix~\ref{sec:limitations}. Token reductions marked $^{*}$ in Table~\ref{tab:eval-results} are significant ($p < 0.05$), confirming that the compression gains are real.
 
\paragraph{Wall-clock latency.}
Table~\ref{tab:latency} reports wall-clock latency (mean seconds per sample) for all models and benchmarks. Evaluations are performed using vLLM on 2$\times$H200 GPUs with greedy decoding. Latency reductions generally track or exceed token count reductions, confirming that the shorter reasoning traces from supertoken adoption translate into practical inference speedups. Averaged across the three models, wall-clock latency drops by ${\sim}10.5\%$ (QwQ-32B $-14.1\%$, Qwen3-30B-A3B $-13.1\%$, DeepSeek-R1-Distill-Llama-70B $-4.4\%$). QwQ-32B shows the most consistent gains, while DeepSeek-R1-Distill-Llama-70B shows variable results across benchmarks.
 
% Latency table — include via \input{table-latency} from appendix
\begin{reusetable}[!htbp]
\centering
\begin{adjustbox}{max width=\textwidth}
\renewcommand{\arraystretch}{1.2}
\begin{tabular}{l cc c}
\toprule
\textbf{Benchmark}
& Baseline Lat.\ (s) & SFT Lat.\ (s) & $\Delta$Lat.\ (\%) ($\downarrow$) \\
\midrule
 
\rowcolor{lightgray}\multicolumn{4}{c}{\textbf{QWQ-32B}} \\
\midrule
AIME'24   & 507 & 466 & -8.1 \\
AIME'25   & 759 & 649 & -14.5 \\
MATH-500  & 1538 & 1252 & -18.6\\
Minerva   & 620 & 545 & -12.1 \\
Olympiad  & 3416 & 2872 & -15.9 \\
\midrule
\textbf{Average} & 1368 & 1157 & \textbf{-13.8} \\
 
\midrule
\rowcolor{lightgray}\multicolumn{4}{c}{\textbf{QWEN-3-30B-A3B}} \\
\midrule
AIME'24   & 2120 & 1844 & \textbf{-13.0} \\
AIME'25   & 2453 & 1922 & \textbf{-21.6} \\
MATH-500  & 2281 & 1882 & -17.5 \\
Minerva   & 2753 & 2402 & -12.7 \\
Olympiad  & 4833 & 4732 & -2.1 \\
\midrule
\textbf{Average} & 2888 & 2556 & \textbf{-13.4} \\
 
\midrule
\rowcolor{lightgray}\multicolumn{4}{c}{\textbf{DeepSeek-R1-Llama-70B-Distill}} \\
\midrule
AIME'24   & 618 & 619.3 & + 0.2 \\
AIME'25   & 1620 & 1483 & -8.5 \\
MATH-500  & 2329 & 1989 & -14.6 \\
Minerva   & 823 & 831.2 & +1.0 \\
Olympiad  & 9110 & 8936 & -1.9 \\
\midrule
\textbf{Average} & 2900 & 2772 & -4.8 \\
\bottomrule
\end{tabular}
\end{adjustbox}
\caption{Wall-clock latency (mean seconds per sample) for baseline and SFT with supertokens, served via SGLang or VLLM on H200 GPUs. The $\Delta$Lat.\ column reports the relative change from Baseline to SFT (supertoken). In the \textbf{Average} rows the latency columns are the mean of the five per-benchmark values, while $\Delta$Lat.\ is the mean of the five per-benchmark percentages rather than the ratio of those averaged latencies; the two differ because absolute runtimes vary by an order of magnitude across benchmarks (e.g.\ OlympiadBench dominates a pooled ratio). Latency reductions exceed or track token count reductions from Table~\ref{tab:eval-results}.}
\label{tab:latency}
\end{reusetable}
 
% --- Supertoken category colors (auto-generated from CATEGORY_COLORS) ---
% These match the colors used in all matplotlib figures.
\definecolor{catbacktracking}{rgb}{0.906, 0.298, 0.235}
\definecolor{cathedging}{rgb}{0.953, 0.612, 0.071}
\definecolor{catverification}{rgb}{0.180, 0.800, 0.443}
\definecolor{catadversative}{rgb}{0.902, 0.494, 0.133}
\definecolor{catstrategyshift}{rgb}{0.608, 0.349, 0.714}
\definecolor{catproblemgrounding}{rgb}{0.204, 0.596, 0.859}
\definecolor{catsequencing}{rgb}{0.102, 0.737, 0.612}
\definecolor{catcontinuation}{rgb}{0.584, 0.647, 0.651}
\definecolor{catnumeric}{rgb}{0.741, 0.765, 0.780}

% Inline color swatch: small rounded square
\newcommand{\catswatch}[1]{%
  \tikz[baseline=-0.6ex]\node[rounded corners=1.5pt, fill=#1,
    minimum width=8pt, minimum height=8pt, inner sep=0pt] {};%
}

% Colored token pill: tinted background + colored border
\newcommand{\cattok}[2]{%
  \tcbox[on line, arc=2pt, outer arc=2pt,
    boxsep=0pt, left=2.5pt, right=2.5pt, top=1.5pt, bottom=1.5pt,
    boxrule=0.5pt, colframe=#1, colback=#1!15,
    fontupper=\ttfamily\scriptsize]{\strut#2}}

\begin{reusetable}[t]
\centering
\small
\renewcommand{\arraystretch}{1.4}
\setlength{\tabcolsep}{5pt}
\begin{tabular}{@{} c l p{4.2cm} p{5.8cm} @{}}
\toprule
 & \multicolumn{1}{c}{\textbf{Category}} & \multicolumn{1}{c}{\textbf{Description}} & \multicolumn{1}{c}{\textbf{Representative Supertokens}} \\
\midrule
\catswatch{catbacktracking} & Backtracking & Self-correction; pausing to reconsider & \cattok{catbacktracking}{Wait, hold on} \cattok{catbacktracking}{But wait} \cattok{catbacktracking}{Wait, no} \\[3pt]
\catswatch{cathedging} & Hedging & Expressing uncertainty or tentativeness & \cattok{cathedging}{, maybe} \cattok{cathedging}{But maybe} \cattok{cathedging}{, so maybe} \\[3pt]
\catswatch{catverification} & Verification & Checking or validating intermediate results & \cattok{catverification}{, right} \cattok{catverification}{, correct} \cattok{catverification}{let's check} \\[3pt]
\catswatch{catadversative} & Counterargument & Introducing a contradiction or counterpoint & \cattok{catadversative}{, but} \cattok{catadversative}{, but the} \cattok{catadversative}{, but we} \cattok{catadversative}{But the} \\[3pt]
\catswatch{catstrategyshift} & Strategy Shift & Changing approach or trying a new method & \cattok{catstrategyshift}{Let's} \cattok{catstrategyshift}{Let me} \cattok{catstrategyshift}{Now, let's} \cattok{catstrategyshift}{Let's try} \\[3pt]
\catswatch{catproblemgrounding} & Problem Ref. & Re-reading or referencing the problem & \cattok{catproblemgrounding}{, the problem says} \cattok{catproblemgrounding}{But the problem} \\[3pt]
\catswatch{catsequencing} & Sequencing & Organizing steps or structural ordering & \cattok{catsequencing}{First,} \cattok{catsequencing}{Similarly,} \cattok{catsequencing}{Given that} \cattok{catsequencing}{Now, the} \\[3pt]
\catswatch{catcontinuation} & Reasoning & Forward-flowing connectives & \cattok{catcontinuation}{, so} \cattok{catcontinuation}{, and} \cattok{catcontinuation}{, which is} \cattok{catcontinuation}{Therefore,} \\[3pt]
\catswatch{catnumeric} & Computation & Digits or single-letter variables & \cattok{catnumeric}{1} \cattok{catnumeric}{2} \cattok{catnumeric}{, x} \cattok{catnumeric}{, n} \cattok{catnumeric}{, k} \\
\bottomrule
\end{tabular}
\caption{Supertoken structural taxonomy. Each merge in the SuperBPE vocabulary is classified into one of nine structural categories based on its decoded text. Representative supertokens are shown for each category.}
\label{tab:supertoken-taxonomy}
\end{reusetable}

\subsection{Structural analysis of reasoning trajectories}
\label{sec:structural}
 
Beyond compression, the learned supertokens provide a compact and interpretable view of recurring reasoning structure. We classify supertokens into the taxonomy shown in Table~\ref{tab:supertoken-taxonomy} using deterministic keyword-based rules applied in priority order to each merge's decoded text (full rule specification in Appendix~\ref{sec:taxonomy-details}). This maps each trace into a sequence of coarse reasoning moves. Although similar categories can be detected by applying string-matching rules directly to raw reasoning traces, the advantage of supertokens is that high-frequency recurring idioms are learned from the model's own trace distribution and emitted as discrete vocabulary items.
 
\begin{figure}[t]
  \centering
  \includegraphics[width=\linewidth]{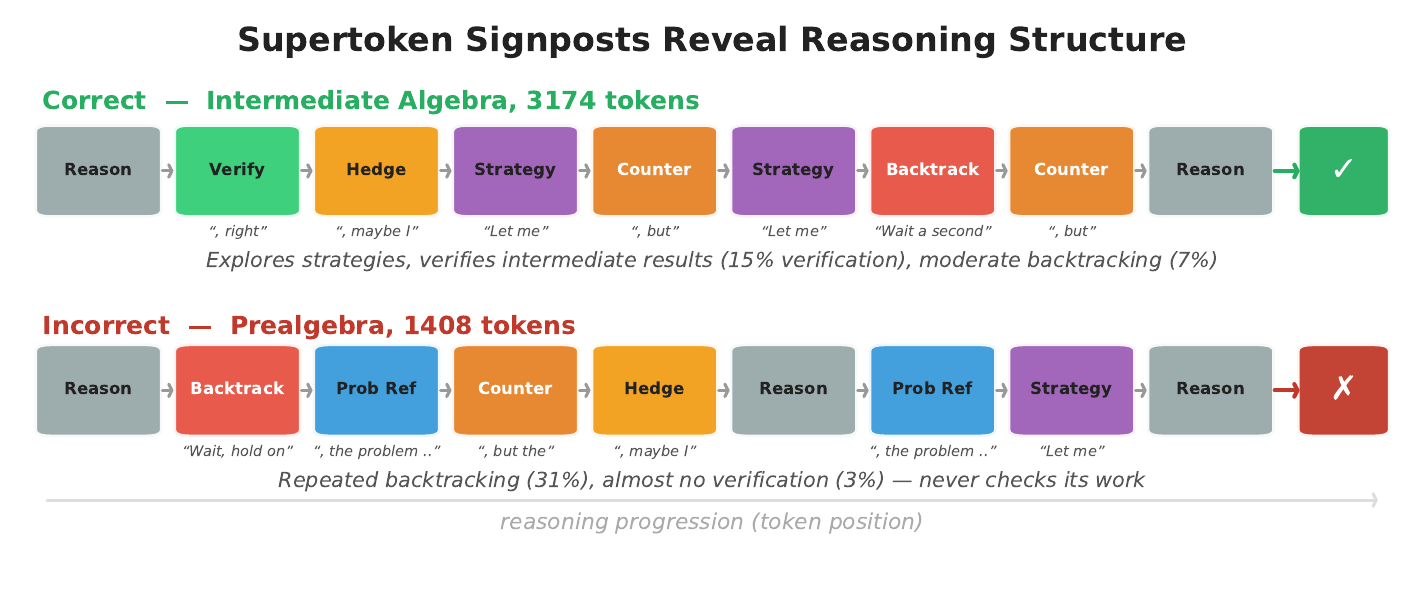}
  \caption{Supertoken signposts reveal reasoning structure. Each block represents a structural reasoning move detected via supertokens. The correct trace (top) shows diverse, balanced reasoning with verification; the incorrect trace (bottom) is dominated by repeated backtracking without verification.}
  \label{fig:schematic-reasoning}
\end{figure}
 
Figure~\ref{fig:schematic-reasoning} shows two reasoning traces as illustrative examples, highlighting the logical flow and key transitions throughout each trajectory. The correct trace exhibits a balanced structure: the model explores strategies, verifies intermediate results, and backtracks sparingly. In contrast, the incorrect trace is dominated by repeated backtracking with little verification, suggesting repeated self-correction without sufficient checking.
 
Aggregating supertoken transitions by answer correctness shows that the structural differences illustrated above extend across the full evaluation set. We report two clusters of transition bigrams that are over-represented in one class relative to the other (Table~\ref{tab:diagnostic-transitions}); these are aggregate over-representations across the evaluation pool, not held-out predictive evaluations of trace correctness. \emph{Incorrect} traces show elevated rates of what we descriptively label \textbf{confusion cycles}: problem-reference$\to$hedging ($2.1\times$ over-represented), counterargument$\to$problem-reference ($2.0\times$), and hedging$\to$hedging ($1.4\times$). \emph{Correct} traces show elevated rates of \textbf{productive recovery}: problem-reference$\to$strategy-shift ($3.0\times$), verification$\to$strategy-shift ($2.1\times$), and reasoning$\to$verification ($1.6\times$). The full transition probability matrix is shown in Appendix~\ref{sec:transition-appendix}.

\begin{reusetable}[t]
\centering
\small
\renewcommand{\arraystretch}{1.3}
\setlength{\tabcolsep}{4pt}
\begin{tabular}{@{} l l c l @{}}
\toprule
\multicolumn{1}{c}{\textbf{From}} & \multicolumn{1}{c}{\textbf{To}} & \multicolumn{1}{c}{\textbf{Ratio}} & \multicolumn{1}{c}{\textbf{Interpretation}} \\
\midrule
\multicolumn{4}{l}{\textit{Problematic transitions} (over-represented in incorrect traces)} \\[3pt]
Sequencing & Sequencing & $3.7\times$ & Repeated reorganization without progress \\
Problem Ref. & Hedging & $2.1\times$ & Re-reads problem $\to$ becomes uncertain \\
Counterargument & Problem Ref. & $2.0\times$ & Contradiction $\to$ re-reads problem (confusion loop) \\
Hedging & Hedging & $1.4\times$ & Sustained unresolved uncertainty \\
Counterargument & Counterargument & $1.3\times$ & Contradictions accumulating unresolved \\
\midrule
\multicolumn{4}{l}{\textit{Productive transitions} (over-represented in correct traces)} \\[3pt]
Problem Ref. & Strategy Shift & $3.0\times$ & Re-reads problem $\to$ devises new plan \\
Verification & Strategy Shift & $2.1\times$ & Checks result $\to$ adapts approach \\
Reasoning & Verification & $1.6\times$ & Forward reasoning $\to$ verifies work \\
Backtracking & Strategy Shift & $1.5\times$ & Recognizes error $\to$ pivots constructively \\
\bottomrule
\end{tabular}
\caption{Bigram supertoken transitions over-represented by trace correctness. \textit{Problematic} bigrams are over-represented in incorrect traces; \textit{productive} bigrams are over-represented in correct traces. Ratios are aggregate over-representations across the evaluation pool; these are descriptive correlational patterns and not held-out predictive evaluations of correctness.}
\label{tab:diagnostic-transitions}
\end{reusetable}

\section{Discussion}
 
Our results suggest that reasoning traces contain a substantial amount of repeated low-entropy scaffolding. By learning vocabulary items for these recurring phrases, supertokens compress the expression of reasoning without removing intermediate steps or replacing them with latent states. This gives a different trade-off from content-level or latent CoT compression methods: the compression ratio is more modest, but the resulting traces remain readable and largely faithful to the original reasoning process. The cost profile also differs: supertoken adoption requires only a single small round of plain SFT, with no reward model, controller, or inference-time routing, and the resulting tokenizer and model are a permanent drop-in change with zero per-request overhead, so the training cost is one-time and amortized over all subsequent inference.

A useful byproduct of this vocabulary-level compression is interpretability. Many learned supertokens correspond to recognizable reasoning idioms such as backtracking, verification, strategy shifts, and hedging, making the compressed trace easier to inspect at a structural level. These analyses do not require supertokens in principle; similar categories can be detected by string matching over raw traces, while supertokens provide a learned vocabulary-level interface for inspecting frequent idioms.
 
The transition patterns we observe are correlational: they describe aggregate over-representations across our evaluation pool, not whether any individual transition is predictive of correctness on a held-out trace. We therefore present them as a descriptive structural analysis only. Whether bigram-level transition features add held-out predictive signal beyond simple baselines such as trace length, and, if so, whether they could in turn serve as early-stopping or reward-shaping signals in RL-based reasoning training, is a natural follow-up that we leave to future work.

Overall, reasoning supertokens provide a simple, model-agnostic way to compress LLM reasoning traces while preserving readability, without pruning steps or replacing them with latent states. Across three model families and five benchmarks, this vocabulary-level approach achieves 8.1\% average compression with accuracy equivalent or inconclusive at $\pm 2$~pp on 14/15 model--benchmark cells (TOST; 3 equivalent, 11 inconclusive), with one non-equivalent degradation on DeepSeek-R1-Distill-Llama-70B (Appendix~\ref{sec:limitations}), while exposing recurring structural patterns in the generated traces. Appendix~\ref{sec:limitations} discusses the limitations of this work: the modest compression ratio relative to content-level methods, the restriction of our evaluation to mathematical reasoning, the dependence on representative traces from the target model, and the accuracy degradation observed on DeepSeek-R1-Distill-Llama-70B.

\bibliography{colm2026_conference}
\bibliographystyle{colm2026_conference}
 
\appendix
 
\section{Limitations}
\label{sec:limitations}
 
Our approach achieves more modest compression (${\sim}$8.1\%) compared to content-level CoT compression methods that report 30--70\% reductions (Table~\ref{tab:method-comparison}). This comparison spans two different reference classes: as a drop-in tokenizer change, our method should be measured against the tokenizer-algorithm literature, where corpus-level gains are typically $1$--$2\%$ (PickyBPE~\citep{chizhov-etal-2024-bpe}: ${\sim}1.1\%$) and $8.1\%$ is large, whereas the $30$--$70\%$ figures come from content-level methods that remove or shorten reasoning steps and report measurable accuracy loss on harder benchmarks. The more modest ratio is a deliberate design choice: by operating at the vocabulary level, we compress the \emph{tokenization} of reasoning rather than its \emph{content}, preserving the complete human-readable trace.

Our method also depends on the availability of representative reasoning traces from the target model, and the learned vocabulary may not transfer cleanly across tokenizer families.
 
\paragraph{Domain coverage.} Our evaluation is restricted to mathematical reasoning. The structural idioms our supertokens capture (e.g., \textit{``let me think''}, \textit{``wait''}, \textit{``verify''}) are largely domain-general reasoning markers, so we expect partial transfer, but we have not validated this and do not claim it. Coding, agentic planning, and scientific or long-form QA are the untested domains and the natural next evaluation. The IFEval results in Appendix~\ref{sec:ifeval-appendix} indicate that general instruction-following is preserved after supertoken SFT, but this does not by itself establish that the compression transfers to non-math reasoning.
 
\paragraph{Accuracy degradation on DeepSeek-R1-Distill-Llama-70B.} Our TOST analysis at $\pm 2$~pp (Appendix~\ref{sec:significance-appendix}) identifies one model--benchmark cell where the accuracy drop is non-equivalent and the 90\% CI excludes 0 in the negative direction (a real accuracy loss): DeepSeek-R1-Distill-Llama-70B on MATH-500 ($\Delta = -3.3$~pp, 90\% CI $[-5.38, -1.22]$). This is also the model on which we observe the largest absolute token compression. Two further DeepSeek-R1-Distill-Llama-70B cells with negative point estimates, OlympiadBench ($\Delta = -1.5$~pp, 90\% CI $[-3.57, +0.57]$) and AIME'25 ($\Delta = -1.1$~pp, 90\% CI $[-11.21, +9.01]$ at $N{=}30$), are statistically inconclusive rather than demonstrated losses. The Qwen-family models show no non-equivalent cells. Cleanly disentangling tokenizer-family effects from model-specific effects requires evaluating a second Llama-family model (e.g., R1-Distill-Llama-8B or Llama-3.3-70B-Instruct), which is the priority follow-up we plan to run.
 
\clearpage
 
\section{Reasoning Supertokens}
\label{sec:supertoken-details}
Table~\ref{tab:token-examples} shows representative supertokens under different filtering rules. The \emph{structural filter} (Section~\ref{sec:vocab}) removes high-frequency patterns unrelated to reasoning (e.g., \textit{``is the''}, \textit{``in the''}) and possessive artifacts starting with \textit{``'s''}, while retaining compositional reasoning phrases such as \textit{``Let's check''} and \textit{``Let's reconsider''}.
 
\begin{reusetable}[H]
\centering
\small
\begin{tabular}{lrr}
\toprule
Token & No filter & Filtered \\
\midrule
\multicolumn{3}{l}{\textit{Kept by both strategies}} \\[2pt]
\tokens{{.},{\spnl}}                          & 261,553 & 261,553 \\
\tokens{{.},{\spnl\spnl}}                     & 254,138 & 254,138 \\
\tokens{{:},{\spnl}}                          & 243,970 & 243,970 \\
\tokens{{,},{\spspace{}the}}                  & 242,352 & 242,352 \\
\tokens{{\spspace{}},{1}}                     & 264,450 & 264,450 \\
\tokens{{,},{\spspace{}then}}                 & 193,242 & 193,242 \\
\tokens{{What},{\spspace{}if}}                & 136,224 & 136,224 \\
\tokens{{So},{\spspace{}the}}                 & 135,171 & 135,171 \\
\midrule
\multicolumn{3}{l}{\textit{Removed by filter}} \\[2pt]
\tokens{{\spspace{}is},{\spspace{}the}}       & 241,551 & -- \\
\tokens{{\spspace{}is},{\spspace{}a}}         & 241,413 & -- \\
\tokens{{\spspace{}in},{\spspace{}the}}       & 231,043 & -- \\
\tokens{{\spspace{}of},{\spspace{}the}}       & 230,200 & -- \\
\tokens{{'s},{\spspace{}check}}               & 219,883 & -- \\
\tokens{{'s},{\spspace{}assume}}              & 165,624 & -- \\
\tokens{{impl},{ies}}                         & 196,004 & -- \\
\tokens{{?},{\spspace{}No},{.},{\spnl}}       & 210,179 & -- \\
\midrule
\multicolumn{3}{l}{\textit{Surfaced by filter}} \\[2pt]
\tokens{{Let},{'s}}                                          & -- & 189,651 \\
\tokens{{Is},{\spspace{}it}}                                 & -- & 156,033 \\
\tokens{{Let},{'s},{\spspace{}check}}                        & -- & 152,862 \\
\tokens{{Is},{\spspace{}it},{\spspace{}possible}}            & -- & 151,487 \\
\tokens{{This},{\spspace{}is}}                               & -- & 147,188 \\
\tokens{{Is},{\spspace{}there}}                              & -- & 142,627 \\
\tokens{{Wait},{,},{\spspace{}if}}                           & -- & 127,788 \\
\tokens{{Let},{'s},{\spspace{}assume}}                       & -- & 124,009 \\
\tokens{{This},{\spspace{}implies}}                          & -- & 105,077 \\
\tokens{{Let},{'s},{\spspace{}reconsider}}                   & -- & 47,570  \\
\tokens{{Let},{'s},{\spspace{}consider}}                     & -- & 45,207  \\
\bottomrule
\multicolumn{3}{l}{\footnotesize Frequencies at 150-merge budget. {--} indicates token absent from that merge table.} \\
\end{tabular}
\caption{Representative super-tokens from unfiltered and filtered BPE.
The filter removes syntactically promiscuous merges and math artifacts, and by admitting apostrophes
surfaces a compositional family of reasoning-specific phrases built on \tokens{{Let},{'s}} as a shared prefix.}
\label{tab:token-examples}
\end{reusetable}
 
\subsection{Exemplary Reasoning Traces}
\label{sec:trace-examples}
\begin{figure}[!htbp]
  \centering
  \includegraphics[width=\linewidth]{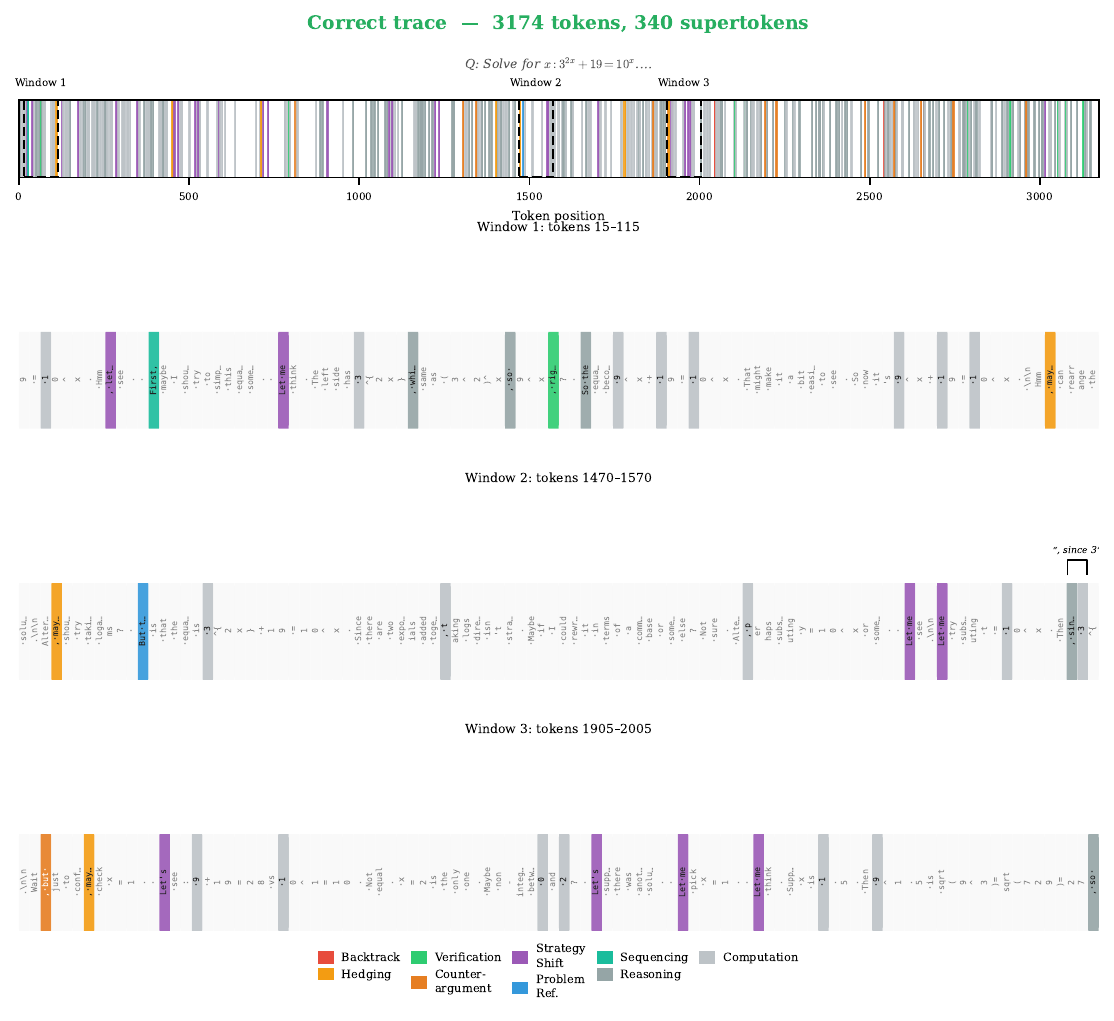}
  \caption{%
    Correct reasoning trace with supertoken categories. Top: ribbon overview showing category distribution across all 3{,}174 tokens. Bottom: three zoomed windows at structurally interesting positions. Colors correspond to the taxonomy in Table~\ref{tab:supertoken-taxonomy}; gray tokens are non-merged (organic content).%
  }
  \label{fig:trace-correct}
\end{figure}

\begin{figure}[!htbp]
  \centering
  \includegraphics[width=\linewidth]{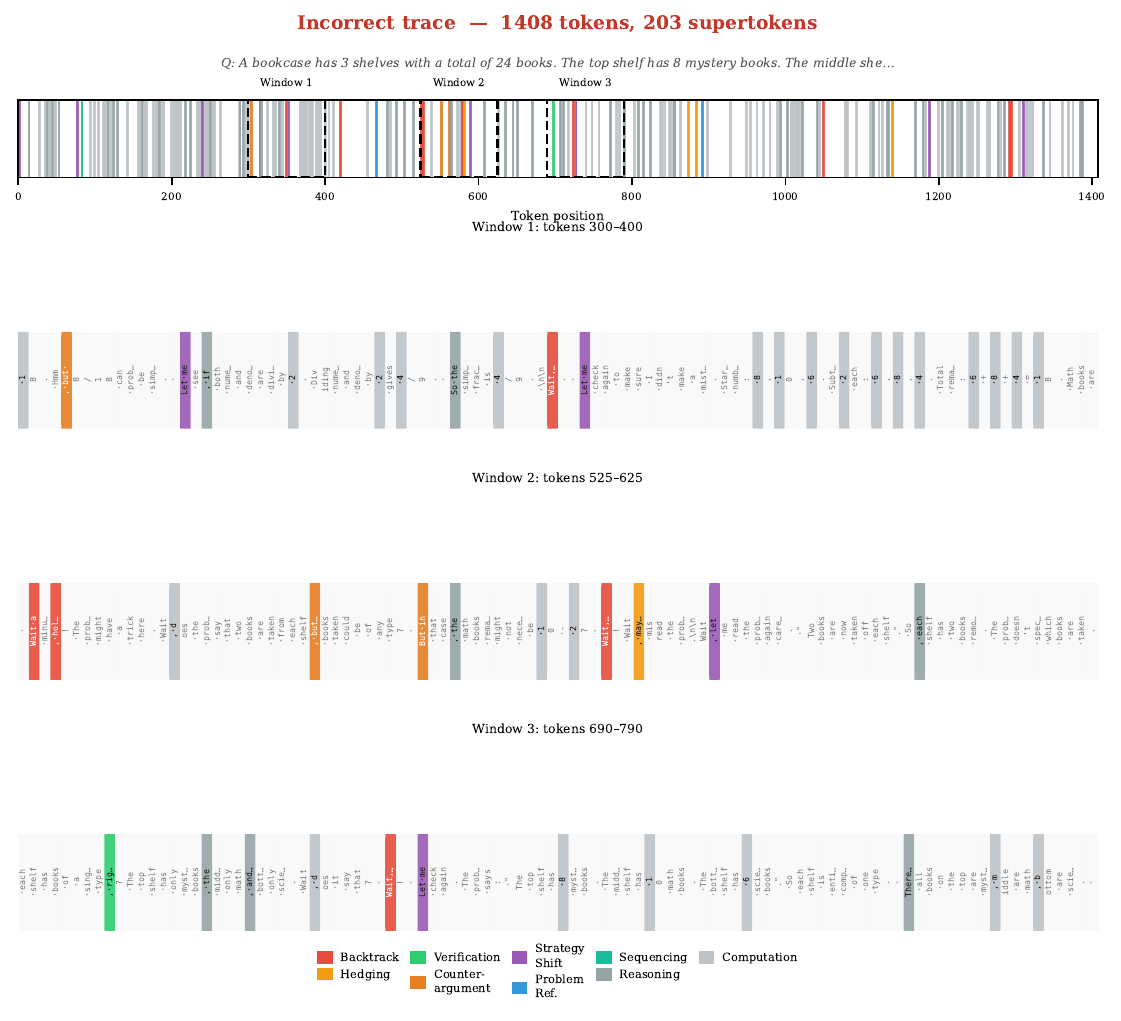}
  \caption{%
    Incorrect reasoning trace with supertoken categories. The ribbon overview reveals a higher density of backtracking (red) and counterargument (orange) tokens compared to the correct trace. Zoomed windows show repeated self-correction cycles without verification, illustrating the confusion patterns identified in Section~\ref{sec:structural}.%
  }
  \label{fig:trace-incorrect}
\end{figure}

Figures~\ref{fig:trace-correct} and~\ref{fig:trace-incorrect} show two complete reasoning traces rendered at the token level, with supertoken merge positions colored by their structural category (Table~\ref{tab:supertoken-taxonomy}). Each figure consists of a compact \emph{ribbon} overview (top) showing the category distribution across the full trace, and three \emph{zoomed windows} that reveal token-level detail at structurally interesting points.
 
The correct trace (Figure~\ref{fig:trace-correct}) exhibits a diverse mix of categories throughout: strategy shifts (purple) and verification (green) appear regularly, interspersed with forward reasoning. In contrast, the incorrect trace (Figure~\ref{fig:trace-incorrect}) is dominated by backtracking (red) and counterarguments (orange) with notably fewer verification events, consistent with the confusion-cycle pattern described in Section~\ref{sec:structural}.
 
\subsection{Taxonomy Construction Details}
\label{sec:taxonomy-details}
The nine-category taxonomy in Table~\ref{tab:supertoken-taxonomy} is applied via a deterministic, rule-based classifier that operates on the decoded text of each supertoken merge. Classification proceeds by evaluating pattern-matching rules in a fixed priority order; the first matching rule determines the category. No machine-learned component or manual per-token annotation is involved: the same script reproducibly assigns categories to any set of merges.
 
\paragraph{Rule summary.}
Table~\ref{tab:taxonomy-rules} lists the keyword or regex pattern that triggers each category, ordered by evaluation priority (highest first). Higher-priority rules ensure that composite phrases are resolved correctly; for example, \textit{``Wait, let's check''} matches \emph{Backtracking} (via the ``wait'' prefix) before it could match \emph{Verification} (via ``check'') or \emph{Strategy Shift} (via ``let's'').
 
\begin{table}[H]
\centering
\small
\renewcommand{\arraystretch}{1.15}
\begin{tabular}{@{} c l l r @{}}
\toprule
\textbf{Pri.} & \textbf{Category} & \textbf{Key patterns (decoded text)} & \textbf{$N$} \\
\midrule
1 & Backtracking     & starts with ``Wait''; contains ``hold on'', ``but actually'' & 16 \\
2 & Hedging          & contains ``maybe'' & 12 \\
3 & Verification     & contains ``check'', ``matches'', ``which is what''; & 12 \\
  &                  & exact match: ``correct'', ``yes'', ``right'', ``good'', ``impossible'' & \\
4 & Problem Ref.     & contains ``problem'' or ``it says'' & 8 \\
5 & Strategy Shift   & starts with ``Let's'', ``Let me'', ``So let'', ``But let'' & 12 \\
6 & Sequencing       & starts with ``First'', ``Similarly'', ``Given that'', ``Now, the'' & 5 \\
7 & Computation      & single digit (\texttt{0}--\texttt{9}) or single-letter variable & 36 \\
  &                  & (excl.\ ``a'' and ``I'' to avoid articles/pronouns) & \\
8 & Counterargument  & starts with ``But~'' or ``, but'' & 25 \\
9 & Reasoning        & comma-led connectives (``, so'', ``, and'', ``, which'', \ldots); & 124 \\
  &                  & ``Therefore''; ``I think''; fallback: any ``, '' + word & \\
\bottomrule
\end{tabular}
\caption{Taxonomy classification rules applied in priority order. $N$ = number of the 250 supertokens assigned to each category. All 250 merges are classified; no merges fall into an ``other'' residual category.}
\label{tab:taxonomy-rules}
\end{table}
 
\paragraph{Coverage and ambiguity.}
Of the 250 supertokens, all are classified by the deterministic rules (0 fall into the residual ``other'' category). Because rules are evaluated in priority order, composite phrases that could match multiple categories (e.g., \textit{``, but maybe the problem''} contains keywords for Counterargument, Hedging, and Problem Reference) are resolved deterministically. In this example, the ``maybe'' rule (priority~2) fires before ``problem'' (priority~4), yielding \emph{Hedging}, reflecting that the dominant pragmatic function of the phrase is expressing uncertainty rather than referencing the problem. We verified the priority ordering by manually inspecting all 25 supertokens that match keywords from two or more categories; in each case the highest-priority rule assigns the pragmatically dominant function.
 
The full classification script and its output (a JSON mapping from token ID to category) are released with our code, enabling exact reproduction of all taxonomy-dependent analyses in \S\ref{sec:structural}.
 
\clearpage
 
\section{Experimental Setup}
\label{sec:experiment-details}
This section describes the compute environment, hyperparameters, and ablation studies used in our experiments.
 
\subsection{Compute and infrastructure}
All SFT runs use 8$\times$H200 GPUs. QwQ-32B and Qwen3-30B-A3B each train for 1 epoch on OpenThoughts3 (${\sim}$114K samples); DeepSeek-R1-Distill-Llama-70B uses the same data and epoch count. Total wall-clock training time ranges from 4--12 hours depending on model size.
 
\subsection{Training hyperparameters}
Training uses a per-device batch size of 1 with 16 gradient accumulation steps (effective batch size 128), a learning rate of $2 \times 10^{-4}$ with cosine decay and 3\% linear warmup, for 1 epoch.
For the partial layer-unfreezing mode used in all main-table runs, we unfreeze the first $N{=}3$ and last $M{=}1$ transformer layers together with the \texttt{embed\_tokens} and \texttt{lm\_head} rows; these values, the learning rate, and the warmup schedule are held fixed across all three models and all five benchmarks, with no per-cell tuning.
We use bfloat16 mixed precision, Flash Attention 2~\citep{dao2024flashattention}, gradient checkpointing, and Fully Sharded Data Parallel (FSDP) with full sharding.
Checkpoints are saved every 100 steps. A separate learning rate can optionally be applied to the embedding/\texttt{lm\_head} parameters independently from the backbone learning rate.
 
\subsection{Minimum training ablation}
To understand the minimum training needed for supertoken adoption, we tested two reduced configurations on Qwen3-30B-A3B: (i)~masking losses from all tokens except merged supertokens; and (ii)~LoRA with ranks $r \in \{1, 2, 4, 8, 16\}$. Results for the LoRA rank sweep are presented in Section~\ref{sec:lora-ablation}.
 
\clearpage
 
\section{Training Ablations}
\label{sec:training-modes}
Here we present the results of the three different training approaches we tested: Embedding-only, LoRA, and partial-layer unfreezing. All three methods freeze the base model's pretrained weights and differ only in which additional parameters receive gradient updates during SFT. Embedding-only trains just the new supertoken rows in \texttt{embed\_tokens} and \texttt{lm\_head}. LoRA adds low-rank adapters ($r{=}16$, $\alpha{=}32$) to attention projections. Partial unfreezing selectively unfreezes the first $N$ and last $M$ transformer layers alongside the embedding rows.
 
\begin{reusetable}[!htbp]
\centering
\renewcommand{\arraystretch}{1.15}
\begin{adjustbox}{max width=\textwidth}
\begin{tabular}{l l ccc}
\toprule
\textbf{Method} & \textbf{Benchmark} & \textbf{Acc (\%)} & \textbf{Avg.\ Tok.\ ($\downarrow$)} & \textbf{Supertoken (\%)} \\
\midrule

%% ---- Baseline (no SFT) ----
\multirow{3}{*}{\textbf{Baseline (no SFT)}}
  & AIME'24  & 77.50 & 14082 & -- \\
  & AIME'25  & 69.20 & 16651 & -- \\
  & MATH-500 & 80.60 & 4283  & -- \\
\midrule

%% ---- Embedding-only ----
\multirow{3}{*}{\textbf{Embedding-only}}
  & AIME'24  & 70 &13899 & 7.2 \\
  & AIME'25  & 57.3 & 17342 & 7.9 \\
  & MATH-500 & 71.4 & 4754 & 10.6 \\
\midrule

%% ---- LoRA ----
\multirow{3}{*}{\textbf{LoRA}}
  & AIME'24  & 80.0 & 14109 & 7.7 \\
  & AIME'25  & 71.7 & 16480 & 7.4 \\
  & MATH-500 & 80.5 & 4427 & 10.6 \\
\midrule

%% ---- Partial unfreezing ----
\multirow{3}{*}{\textbf{Partial Unfreezing}}
  & AIME'24  & 77.7  & 13160 & 7.6  \\
  & AIME'25  & 72.7  & 14880 & 7.7  \\
  & MATH-500 & 80.5  & 4183  & 10.7 \\
\bottomrule
\end{tabular}
\end{adjustbox}
\caption{Comparison of SFT training strategies on QWQ-32B. Partial unfreezing achieves the best compression while maintaining accuracy.}
\label{tab:training-ablation}
\end{reusetable}

Comparing training modes (Table~\ref{tab:training-ablation}), we find that LoRA achieves supertoken adoption but does not yield compression, whereas partial layer unfreezing produces both. This suggests that learning to \emph{use} supertokens and learning to \emph{compress with} them are distinct processes that require different degrees of model flexibility.
 
\subsection{LoRA Rank Ablation}
\label{sec:lora-ablation}
To isolate the effect of adapter capacity on supertoken adoption and compression, we sweep LoRA rank $r \in \{1, 2, 4, 8, 16\}$ on Qwen3-30B-A3B (Table~\ref{tab:lora-ablation-qwen3}). Across all ranks, supertoken adoption is consistent (${\sim}$10--13\%), and accuracy remains robust. Crucially, none of the LoRA configurations produce meaningful compression (token counts remain at or above baseline levels), reinforcing the observation above that adoption and compression are separate phenomena requiring different degrees of model flexibility.
 
\begin{reusetable}[!htbp]
\centering
\renewcommand{\arraystretch}{1.15}
\begin{adjustbox}{max width=\textwidth}
\begin{tabular}{l l ccc}
\toprule
\textbf{Method} & \textbf{Benchmark} & \textbf{Acc (\%)} & \textbf{Avg.\ Tok.\ ($\downarrow$)} & \textbf{Supertoken (\%)} \\
\midrule

%% ---- LoRA r=1 ----
\multirow{3}{*}{\textbf{LoRA $r{=}1$}}
  & AIME'24  & 93.3 & 17695 & 10.3 \\
  & AIME'25  & 83.3 & 24120 & 10.7 \\
  & MATH-500 & 83.0 & 5832  & 12.9 \\
\midrule

%% ---- LoRA r=2 ----
\multirow{3}{*}{\textbf{LoRA $r{=}2$}}
  & AIME'24  & 93.3 & 18589 & 10.3 \\
  & AIME'25  & 83.3 & 22756 & 10.6 \\
  & MATH-500 & 83.6 & 6024  & 12.8 \\
\midrule

%% ---- LoRA r=4 ----
\multirow{3}{*}{\textbf{LoRA $r{=}4$}}
  & AIME'24  & 86.7 & 18517 & 10.3 \\
  & AIME'25  & 86.7 & 23729 & 10.4 \\
  & MATH-500 & 81.8 & 5920  & 12.8 \\
\midrule

%% ---- LoRA r=8 ----
\multirow{3}{*}{\textbf{LoRA $r{=}8$}}
  & AIME'24  & 86.7 & 18170 & 10.3 \\
  & AIME'25  & 76.7 & 24371 & 10.8 \\
  & MATH-500 & 82.8 & 5814  & 12.8 \\
\midrule

%% ---- LoRA r=16 ----
\multirow{3}{*}{\textbf{LoRA $r{=}16$}}
  & AIME'24  & 93.3 & 17655 & 10.2 \\
  & AIME'25  & 83.4 & 24055 & 10.4 \\
  & MATH-500 & 81.4 & 6024  & 12.8 \\
\bottomrule
\end{tabular}
\end{adjustbox}
\caption{LoRA rank ablation on Qwen3-30B-A3B. All runs use $\alpha{=}32$, targets $\{$q,k,v,o$\}$\_proj, with embed/lm\_head as modules\_to\_save. Supertoken usage is consistent across ranks (${\sim}10$--$13\%$); accuracy is robust to rank choice.}
\label{tab:lora-ablation-qwen3}
\end{reusetable}

\clearpage
 
\section{Merge Selection and Compression}
\label{sec:merge-compression-appendix}
 
\begin{figure}[!htbp]
  \centering
  \includegraphics[width=\linewidth]{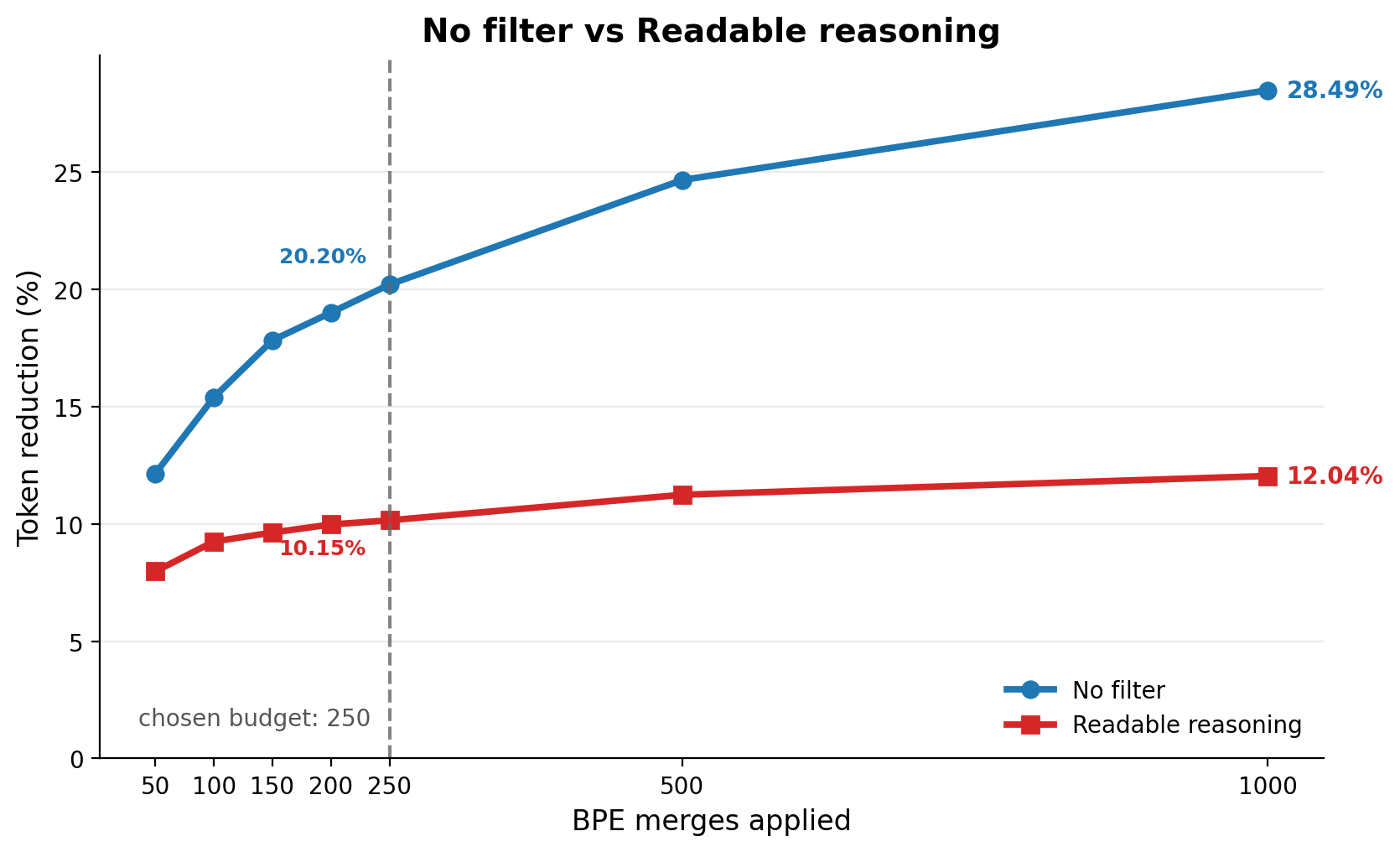}
  \caption{Token compression rate based on merge selection rules and number of merges included.}
  \label{fig:merge-compression}
\end{figure}
 
\subsection{Filter-strength operating points on QwQ-32B}
\label{sec:filter-strength-appendix}
The structural filter used in the main paper (Section~\ref{sec:vocab}) was chosen to preserve interpretability of the resulting supertokens: every retained merge corresponds to a recognizable reasoning idiom usable for the structural analysis in Section~\ref{sec:structural}. The same vocabulary-expansion and SFT pipeline, however, supports a continuum of filter strengths, and the headline compression ratio is one point on a compression--interpretability frontier rather than a property of the framework. To make this concrete, we re-run the QwQ-32B pipeline with a single-rule \emph{light filter} that discards only merges whose decoded text contains no alphabetic character (pure punctuation, pure digit, or whitespace-only chunks) and retains $K{=}1000$ raw merges, yielding 561 unique vocabulary entries after cascade-pruning. The training recipe (3-layer first/1-layer last unfreeze, embed and lm\_head trainable, $\textsc{lr}{=}2{\times}10^{-5}$, OpenThoughts3-1.2M, 8$\times$H200) is otherwise identical to the main run. Removing the filter entirely on the same $K{=}1000$ vocabulary raises the \emph{offline} ceiling to $17.7\%$ but the resulting model fails to convert offline savings into inference savings: at early checkpoints we observe a $25\%$ \emph{increase} in generated length and a macro-accuracy collapse from $0.63$ to $0.40$, because probability mass splits between the supertoken and base-token spellings of the same low-information chunk (e.g.\ ``\texttt{ 0}'' vs.\ ``\texttt{ }''~+~``\texttt{0}'') without the model committing to either. The light-filter configuration in Table~\ref{tab:lightfilt-comparison} is therefore the empirical edge of the useful frontier: it delivers $\sim$67\% more relative token reduction than the paper's strong filter at matched accuracy, and approximates its offline ceiling of $11.8\%$ closely in actual inference.
 
% Filter-strength comparison on QwQ-32B (paper's strong filter, K=250 vs. light filter, K=1000)
% Include via \input{table-lightfilt-comparison} from the main .tex file.
% Requires globally: booktabs, adjustbox, xcolor (posgreen defined in preamble).

\begin{reusetable}[!htbp]
\centering
\renewcommand{\arraystretch}{1.2}
\begin{adjustbox}{max width=\textwidth}
\begin{tabular}{l cc ccc ccc}
\toprule
& \multicolumn{2}{c}{\textbf{QwQ-32B base}\textsuperscript{\dag}}
& \multicolumn{3}{c}{\textbf{Strong filter, $K{=}250$}}
& \multicolumn{3}{c}{\textbf{Light filter, $K{=}1000$}} \\
\cmidrule(lr){2-3} \cmidrule(lr){4-6} \cmidrule(lr){7-9}
\textbf{Benchmark}
 & Acc (\%) & Tok.
 & Acc (\%) & Tok. & Sup.\ (\%)
 & Acc (\%) & Tok. & Sup.\ (\%) \\
\midrule
AIME'24        & 77.3 & 14260 & 73.3 & 13119 &  7.6 & 75.0 & 13167 &  9.4 \\
AIME'25 I      & 64.0 & 16554 & 60.0 & 15582 &  7.9 & 65.3 & 14693 &  9.9 \\
AIME'25 II     & 69.3 & 17445 & 73.3 & 15367 &  7.4 & 62.7 & 15809 &  8.9 \\
MATH-500       & 80.9 &  4387 & 80.6 &  4164 & 10.7 & 80.3 &  4206 & 11.3 \\
Minerva        & 32.6 &  5795 & 33.5 &  5418 &  4.1 & 33.6 &  5352 & 10.0 \\
OlympiadBench  & 56.6 &  9554 & 56.5 &  9256 &  4.5 & 55.8 &  9001 & 10.3 \\
\midrule
\textbf{Macro avg.}
 & \textbf{63.5} & \textbf{11333}
 & \textbf{62.9} & \textbf{10484} & \textbf{7.0}
 & \textbf{62.1} & \textbf{10371} & \textbf{10.0} \\
\textbf{$\Delta$ vs.\ base}
 & -- & --
 & $-0.6$~pp & \textcolor{posgreen}{\textbf{$-7.5\%$}} & --
 & $-1.4$~pp & \textcolor{posgreen}{\textbf{$-8.5\%$}} & -- \\
\midrule
\textbf{Offline ceiling}\textsuperscript{\ddag}
 & -- & --
 & -- & $-7.6\%$ & --
 & -- & $-11.8\%$ & -- \\
\bottomrule
\end{tabular}
\end{adjustbox}
\caption{Effect of filter strength on QwQ-32B reasoning compression. The \emph{strong filter} ($K{=}250$, four surface-form rules) is the configuration used in the main paper and prioritizes interpretability of the resulting supertokens (Section~\ref{sec:vocab}). The \emph{light filter} ($K{=}1000$) keeps the same vocabulary-expansion and SFT recipe but drops a single rule -- discarding only merges whose decoded text contains no alphabetic character (pure punctuation, pure digit, or whitespace-only chunks such as ``\texttt{ 0}'', ``\texttt{ (1}'', ``\texttt{).}'') -- yielding 561 unique vocabulary entries after cascade-pruning (mean merged length 10.6 characters). The light filter delivers $1$~pp more raw token reduction at the cost of $0.8$~pp macro accuracy, with the accuracy difference concentrated on AIME at $N{=}30$ where per-question variance dominates (cf.\ Table~\ref{tab:tost}). Both configurations stay close to their offline compression ceilings on a 200-sample OpenThoughts3 slice ($-7.6\%$ and $-11.8\%$ respectively), confirming that filter strength is the dominant lever between the two operating points. \textsuperscript{\dag}Base = QwQ-32B with identical SFT on OpenThoughts3 \emph{without} vocabulary expansion (matched control, checkpoint 400, 5 replicas per task). The light-filter row reports the step-800 checkpoint averaged over 4--6 replicas per task; the strong-filter row reuses the existing paper run (step-1200, 1--2 replicas per task).}
\label{tab:lightfilt-comparison}
\end{reusetable}

\clearpage
 
\section{Entropy and Compression Ceiling Analyses}
\label{sec:entropy-compression-appendix}
 
\begin{figure}[!htbp]
  \centering
  \includegraphics[width=\linewidth]{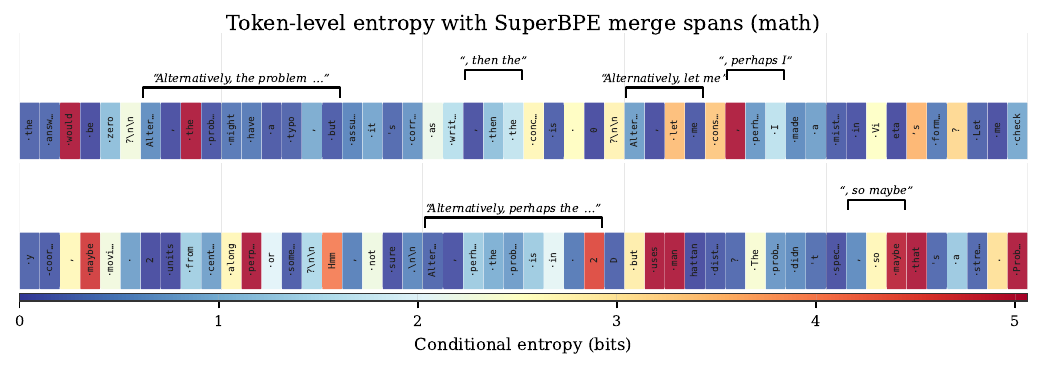}
  \caption{%
    Token-level conditional entropy for two windows from QwQ-32B reasoning traces (blue~=~low, red~=~high). Black brackets mark supertoken merge spans ($\geq$3 base tokens). Within bracketed spans, continuation tokens are predominantly low-entropy, confirming that structural phrases are near-deterministic once begun (continuation mean: 1.03~bits vs.\ non-merge mean: 1.27~bits).%
  }
  \label{fig:entropy-heatmap}
\end{figure}
 
\subsection{Compression Ceiling Analysis}
\label{sec:compression-ceiling-appendix}
\begin{figure}[!htbp]
  \centering
  \includegraphics[width=\linewidth]{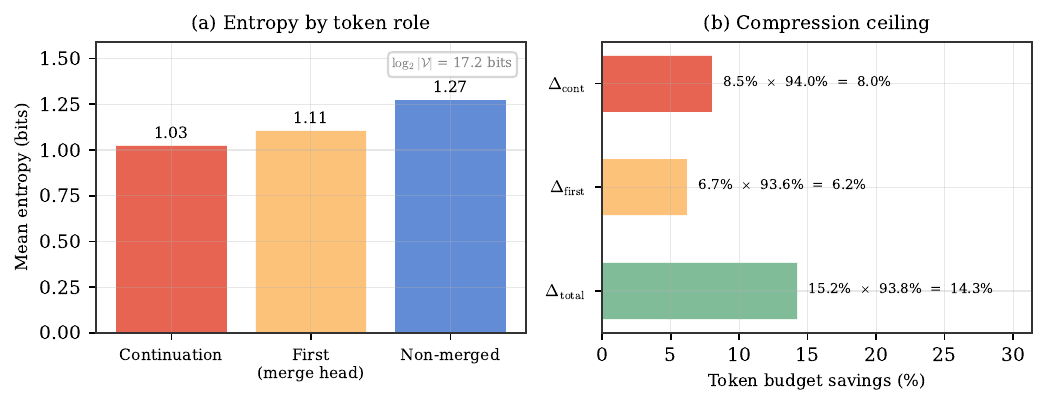}
  \caption{Theoretical compression ceiling.
  (a)~Mean entropy by token role (continuation: 1.03, merge heads: 1.11, non-merged: 1.27~bits), all well below the vocabulary maximum of 17.2~bits.
  (b)~Compression ceiling decomposition ($\Delta = \rho \times (1 - \bar{H}/\log_2|\mathcal{V}|)$). Compressibility is ${\sim}94\%$ across all roles; the combined ceiling $\Delta = 14.3\%$.}
  \label{fig:compression-ceiling}
\end{figure}

Applying Equation~\ref{eq:compression_ceiling} with $\rho = 15.2\%$ and
$h_\mathcal{M} / \log|\mathcal{V}| \approx 0.06$, the compression ceiling is
$\Delta = 14.3\%$, with $\Delta/\rho = 0.94$.
Merged tokens use only 6\% of the vocabulary's information capacity, so nearly all can be absorbed.
Continuation tokens contribute $\Delta_\mathrm{cont} = 8.0\%$ and merge heads $\Delta_\mathrm{first} = 6.2\%$; the effect holds across domains (math $14.9\%$, code $13.6\%$, science $11.5\%$).
Figure~\ref{fig:compression-ceiling} visualizes this decomposition.
 
\subsection{Cross-Model Entropy Validation}
\label{sec:crossmodel-entropy}
The entropy analysis in Section~\ref{sec:entropy-validation} uses QwQ-32B as both generator and scorer, raising the question of whether the observed structural/organic entropy gap is an artifact of a single model's idiosyncrasies. To address this, we re-score the same QwQ-32B reasoning traces using Qwen3-30B-A3B as the scorer model. Both models share the Qwen tokenizer family, allowing direct token-level comparison.
 
Table~\ref{tab:crossmodel-entropy} reports the results. Qwen3-30B-A3B assigns higher entropy across all token roles, as expected for a model that did not generate the traces. Crucially, the entropy ordering is preserved: continuation tokens remain the lowest-entropy role (1.36~bits), followed by merge heads (1.47~bits), with non-merged tokens highest (1.98~bits). The structural/organic gap (non-merged $-$ continuation) actually \emph{widens} from 0.26~bits under self-scoring to 0.62~bits under cross-scoring, indicating that the low entropy of structural tokens is not an artifact of the generating model's own distribution but reflects genuinely predictable patterns in the text. Token-level correlations between the two scorers are high ($r = 0.77$--$0.84$ across roles), confirming that the entropy landscape is consistent across models.
 
% Cross-model entropy validation — include via \input{table-crossmodel-entropy}
\begin{reusetable}[!htbp]
\centering
\renewcommand{\arraystretch}{1.2}
\begin{tabular}{l cc c c}
\toprule
\textbf{Token role} & \textbf{QwQ-32B} & \textbf{Qwen3-30B-A3B} & \textbf{$\Delta$} & \textbf{Fraction} \\
 & (self-scored) & (cross-scored) & & \\
\midrule
Non-merged    & 1.27 & 1.98 & +0.71 & 84.9\% \\
Merge head    & 1.13 & 1.47 & +0.34 &  7.0\% \\
Continuation  & 1.01 & 1.36 & +0.35 &  8.2\% \\
\midrule
Merged (all)  & 1.06 & 1.41 & +0.35 & 15.1\% \\
\midrule
\textbf{Gap} (non-merged $-$ continuation) & \textbf{0.26} & \textbf{0.62} & & \\
\bottomrule
\end{tabular}
\caption{Mean conditional entropy (bits) by token role for QwQ-32B reasoning traces (2K samples, 22.9M tokens), scored by the generating model (QwQ-32B, self) vs.\ a different model (Qwen3-30B-A3B, cross). The structural/organic entropy gap (bottom row) persists and widens under cross-model scoring, confirming the pattern is not an artifact of a single model's distribution.}
\label{tab:crossmodel-entropy}
\end{reusetable}

\clearpage
 
\section{Category Transition Analysis}
\label{sec:transition-appendix}
 
\begin{figure}[!htbp]
  \centering
  \includegraphics[width=0.85\linewidth]{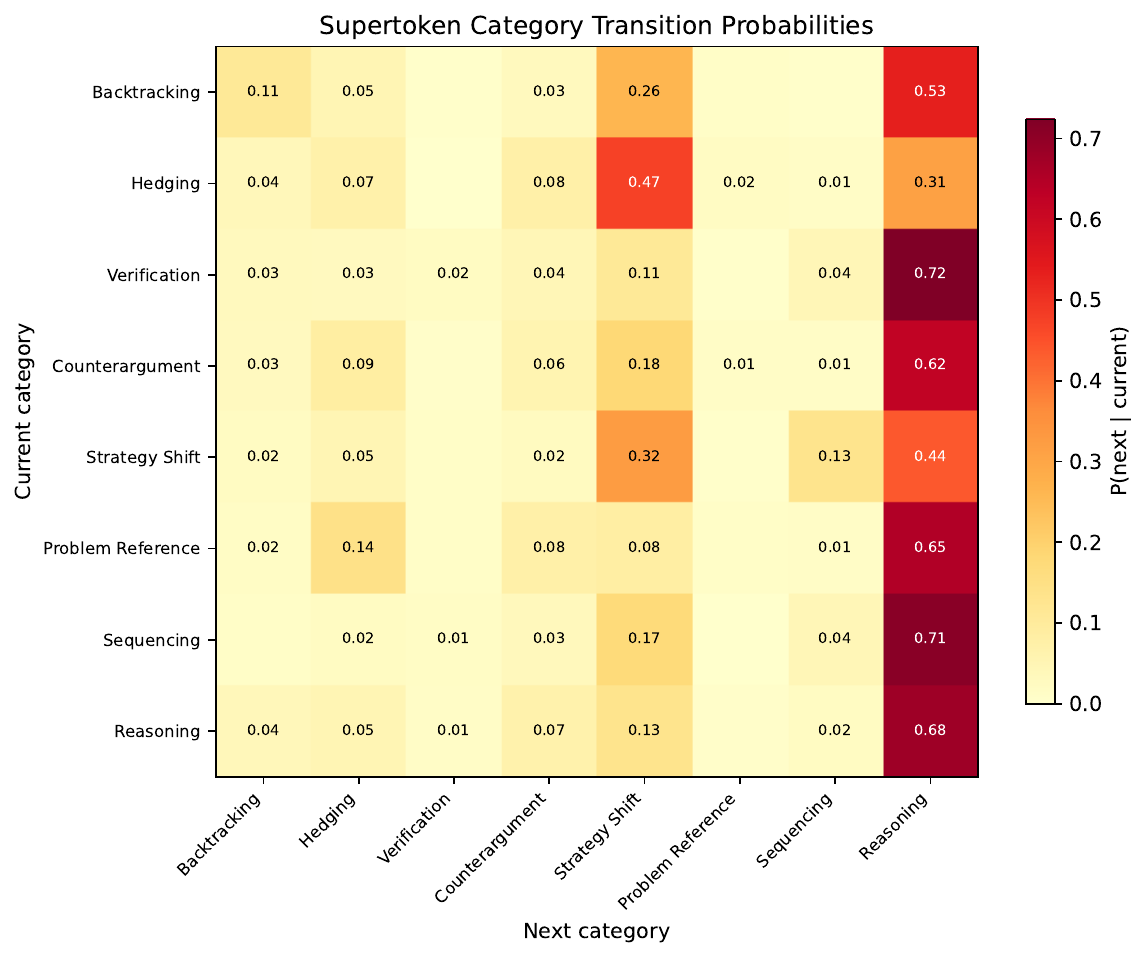}
  \caption{Category transition probabilities aggregated across all evaluation traces. Each cell shows the probability of transitioning from the row category to the column category. High self-transition rates along the diagonal reflect sustained forward computation, while off-diagonal hotspots expose characteristic reasoning patterns such as Backtracking followed by Reasoning (recovery) and Verification followed by Strategy Shift (adaptive re-planning).}
  \label{fig:transition-heatmap}
\end{figure}
 
\clearpage
 
\section{General Instruction-Following Evaluation}
\label{sec:ifeval-appendix}
A potential concern with vocabulary expansion and fine-tuning is that it could degrade the model's general instruction-following ability. To test this, we evaluate all three model families on IFEval~\citep{zhou2023ifeval}, a benchmark that measures compliance with explicit formatting and content constraints (e.g., ``write exactly 3 paragraphs'', ``include the keyword X''). Table~\ref{tab:ifeval} reports the strict prompt-level and instruction-level accuracies before and after supertoken SFT.
 
\begin{reusetable}[!htbp]
\begin{adjustbox}{max width=\textwidth}
\renewcommand{\arraystretch}{1.2}
\centering
\begin{tabular}{l cc cc }
\toprule
& \multicolumn{2}{c}{\textbf{Baseline}} & \multicolumn{2}{c}{\textbf{SFT}}  \\
\cmidrule(lr){2-3} \cmidrule(lr){4-5}
\textbf{Model} & Prompt (\%) & Inst.\ (\%) & Prompt (\%) & Inst.\ (\%)  \\
\midrule
QwQ-32B                        & 81.3 & 87.8 & 79.8 &88.0  \\
Qwen3-30B-A3B                  & 87.2 & 91.5 & 86.5 & 90.3 \\
DeepSeek-R1-Distill-Llama-70B  & 86.3 & 80.0 & 85.9 & 80.2 \\
\bottomrule
\end{tabular}
\end{adjustbox}
\caption{IFEval results before and after supertoken fine-tuning. Prompt-level and instruction-level accuracy remain stable, confirming that vocabulary expansion and SFT do not degrade general instruction-following ability.}
\label{tab:ifeval}
\end{reusetable}

\clearpage
 
\section{Statistical Significance Analysis}
\label{sec:significance-appendix}
We test \emph{equivalence} of accuracy under SFT with supertokens via a two one-sided test (TOST), and report paired confidence intervals for completion-token deltas.
 
\paragraph{TOST equivalence on accuracy.}
Failure-to-reject the null of zero accuracy difference is not equivalence; to make the accuracy-preservation claim explicit, we apply the TOST procedure~\citep{schuirmann1987tost} at a single equivalence margin of $\pm 2$~pp (a strict, small-effect threshold on accuracy). For each (model, benchmark) cell we compute per-question accuracy averaged across the 5 evaluation runs $\bar{p}_q^{\mathrm{cond}} = \tfrac{1}{5}\sum_r \mathbb{1}[\mathrm{correct}_{q,r}^{\mathrm{cond}}]$ for $\mathrm{cond} \in \{\mathrm{base}, \mathrm{sft}\}$, form $N$ paired differences $d_q = \bar{p}_q^{\mathrm{sft}} - \bar{p}_q^{\mathrm{base}}$, and apply a paired $t$-test with df $= N - 1$. The point estimate is anchored to the paper's Table~\ref{tab:eval-results} reported $\Delta$Acc; the standard error is computed from the per-question paired differences. Equivalence holds when the 90\% CI of $\bar{d}$ is contained in $[-2,\,+2]$~pp (Schuirmann TOST, $\alpha = 0.05$). Each cell receives one of three decisions: \textbf{PASS} (90\% CI $\subset [-2, +2]$~pp); \textit{inconclusive} (90\% CI straddles a margin bound -- typically under-powered, predominantly AIME with $N{=}30$); \textbf{FAIL} (90\% CI excludes 0 in the negative direction, indicating a real accuracy loss).
 
\paragraph{Completion token CIs.}
Raw per-question token counts have very high variance (problems range from easy to extremely hard), but because each question is answered by both the baseline and SFT model, per-sample pairing cancels out question-level difficulty. The paired-difference standard deviation is typically 20--40\% of the raw within-condition standard deviation. We estimate the paired-difference SD as $\sigma_{\mathrm{paired}} \approx 0.24 \times \bar{T}$, where $\bar{T}$ is the mean token count across conditions, reflecting ${\sim}$30\% of a raw SD that is ${\sim}$80\% of the mean. The 95\% CI is then $\Delta T \pm 1.96 \cdot \sigma_{\mathrm{paired}} / \sqrt{N}$.
 
\paragraph{Results.}
Table~\ref{tab:tost} reports the per-cell TOST decisions; Table~\ref{tab:significance} reports the completion-token CIs.
For accuracy: \textbf{3/15} cells pass equivalence at the $\pm 2$~pp margin (QwQ-32B/MATH-500, QwQ-32B/OlympiadBench and Qwen3-30B-A3B/MATH-500); \textbf{11/15} are inconclusive (90\% CI straddles a margin bound, predominantly AIME at $N{=}30$ and Minerva/OlympiadBench cells with small positive or negative drifts); and \textbf{1/15} fails equivalence with a 90\% CI excluding 0 in the negative direction: DeepSeek-R1-Distill-Llama-70B on MATH-500 ($\Delta = -3.3$~pp, 90\% CI $[-5.38, -1.22]$), discussed in Appendix~\ref{sec:limitations}.
For completion tokens, 9 of 15 reductions are statistically significant (CI excludes zero), concentrated on the larger benchmarks (MATH-500, OlympiadBench, Minerva) where sample sizes provide sufficient power. On AIME ($N{=}30$), reductions are directionally consistent but often do not reach significance, with the exception of the large compressions for DeepSeek-R1-Distill-Llama-70B ($-17\%$) which are significant even at $N{=}30$.
 
\begin{table}[H]
\centering
\small
\renewcommand{\arraystretch}{1.1}
\begin{adjustbox}{max width=\textwidth}
\begin{tabular}{ll r r c l}
\toprule
\textbf{Model} & \textbf{Benchmark} & {$\bar{d}$ (pp)} & {90\% CI} & {TOST $\pm 2$} & {Decision} \\
\midrule
\multirow{5}{*}{QwQ-32B}
 & AIME'24       & $+0.20$ & $[-5.17,\;+5.57]$  & \ding{55} & \textit{inconclusive} \\
 & AIME'25       & $+3.50$ & $[-2.08,\;+9.08]$  & \ding{55} & \textit{inconclusive} \\
 & MATH-500      & $-0.10$ & $[-1.01,\;+0.81]$  & \ding{51} & PASS \\
 & Minerva       & $-0.50$ & $[-2.21,\;+1.21]$  & \ding{55} & \textit{inconclusive} \\
 & OlympiadBench & $+0.70$ & $[-0.30,\;+1.70]$  & \ding{51} & PASS \\
\midrule
\multirow{5}{*}{Qwen3-30B-A3B}
 & AIME'24       & $-1.00$ & $[-5.10,\;+3.10]$  & \ding{55} & \textit{inconclusive} \\
 & AIME'25       & $+0.10$ & $[-3.89,\;+4.09]$  & \ding{55} & \textit{inconclusive} \\
 & MATH-500      & $+1.10$ & $[+0.21,\;+1.99]$  & \ding{51} & PASS \\
 & Minerva       & $-0.40$ & $[-2.53,\;+1.73]$  & \ding{55} & \textit{inconclusive} \\
 & OlympiadBench & $+1.50$ & $[-0.13,\;+3.13]$  & \ding{55} & \textit{inconclusive} \\
\midrule
\multirow{5}{*}{\shortstack[l]{DeepSeek-R1-\\Distill-Llama-70B}}
 & AIME'24       & $+3.50$ & $[-4.62,\;+11.62]$ & \ding{55} & \textit{inconclusive} \\
 & AIME'25       & $-1.10$ & $[-11.21,\;+9.01]$ & \ding{55} & \textit{inconclusive} \\
 & MATH-500      & $-3.30$ & $[-5.38,\;-1.22]$  & \ding{55} & \textbf{FAIL} \\
 & Minerva       & $-0.20$ & $[-2.57,\;+2.17]$  & \ding{55} & \textit{inconclusive} \\
 & OlympiadBench & $-1.50$ & $[-3.57,\;+0.57]$  & \ding{55} & \textit{inconclusive} \\
\bottomrule
\end{tabular}
\end{adjustbox}
\caption{TOST equivalence analysis on accuracy at a single equivalence margin of $\pm 2$~pp. For each (model, benchmark) we compute per-question accuracy averaged across the 5 evaluation runs and apply a paired $t$-test on the differences (df $= N-1$). Equivalence holds when the 90\% CI of $\bar{d}$ is contained in $[-2,\,+2]$~pp. Decision: \textbf{PASS} = 90\% CI $\subset [-2, +2]$~pp; \textit{inconclusive} = 90\% CI straddles a margin bound (typically under-powered; AIME and small-drift cells); \textbf{FAIL} = 90\% CI excludes 0 in the negative direction (a real accuracy loss is detected). Of 15 cells: \textbf{3 PASS, 11 inconclusive, 1 FAIL}.}
\label{tab:tost}
\end{table}
 
\begin{table}[H]
\centering
\small
\renewcommand{\arraystretch}{1.1}
\begin{adjustbox}{max width=\textwidth}
\begin{tabular}{ll rrc}
\toprule
\textbf{Model} & \textbf{Benchmark} & {$\Delta$Tok} & {95\% CI} & {Sig.} \\
\midrule
\multirow{5}{*}{QwQ-32B}
 & AIME'24       & $-922$  & $[-2092,\;+248]$   & \\
 & AIME'25       & $-1771$ & $[-3125,\;-417]$   & \ding{51} \\
 & MATH-500      & $-258$  & $[-349,\;-167]$    & \ding{51} \\
 & Minerva       & $+94$   & $[-59,\;+247]$     & \\
 & OlympiadBench & $-328$  & $[-499,\;-157]$    & \ding{51} \\
\midrule
\multirow{5}{*}{Qwen3-30B-A3B}
 & AIME'24       & $-1059$ & $[-2414,\;+296]$   & \\
 & AIME'25       & $-974$  & $[-2606,\;+658]$   & \\
 & MATH-500      & $-639$  & $[-748,\;-530]$    & \ding{51} \\
 & Minerva       & $-24$   & $[-126,\;+78]$     & \\
 & OlympiadBench & $-937$  & $[-1170,\;-704]$   & \ding{51} \\
\midrule
\multirow{5}{*}{\shortstack[l]{DeepSeek-R1-\\Distill-Llama-70B}}
 & AIME'24       & $-1540$ & $[-2252,\;-828]$   & \ding{51} \\
 & AIME'25       & $-1942$ & $[-2832,\;-1052]$  & \ding{51} \\
 & MATH-500      & $-160$  & $[-215,\;-105]$    & \ding{51} \\
 & Minerva       & $-113$  & $[-204,\;-22]$     & \ding{51} \\
 & OlympiadBench & $-3$    & $[-118,\;+112]$    & \\
\bottomrule
\end{tabular}
\end{adjustbox}
\caption{Paired 95\% confidence intervals for completion-token deltas. Paired-difference SD estimated as $\sigma_{\mathrm{paired}} \approx 0.24 \times \bar{T}$. \ding{51} marks token reductions whose CI excludes zero. Reductions are significant on larger benchmarks; AIME ($N{=}30$) lacks power except for the large DS-R1-Llama compressions.}
\label{tab:significance}
\end{table}
 
\end{document}